\documentclass[journal,10pt,twoside]{IEEEtran}
\usepackage{verbatim}
\usepackage[pdftex]{graphicx}
\usepackage{epsfig,subfigure}
\usepackage{amsmath,amssymb,amsfonts,bm,amscd} 
\usepackage{algpseudocode}
\usepackage{mathtools} % <-- Add this line
\usepackage{mathrsfs}
\usepackage{fancyhdr}
\usepackage{algorithm} 
\usepackage{psfrag}
\usepackage{cite}
\usepackage{makecell}
\usepackage{url}
\usepackage{float}
\usepackage{soul}
\usepackage{color}
\usepackage{multirow}
\usepackage{booktabs}
\usepackage{etoolbox} 

\usepackage{color, xcolor}

\newtheorem{theorem}{Theorem}[section]

\newtheorem{definition}{Definition}

\newtheorem{assumption}{Assumption}
\newtheorem{remark}{Remark}

\newcommand{\rr}{\mathbb{R}}

\newcommand{\mobility}[0]{ ${e}_{\text{mae}}$ ($\text{m/s}$)
                    }

\newcommand{\safety}[0]
                    { ${\mathcal{P}}_{s}$  ($\text{\%}$)
                    &  ${\mathcal{S}}_{\text{ma}}$  ($\text{m}$)}
\newcommand{\safetyngsim}[0]
                    { ${\mathcal{S}}_{\text{ma}}$  ($\text{m}$)}
 \newcommand{\safetymerge}[0]
                    {   ${\mathcal{S}}_{\text{lon, ma}}$  ($\text{m}$)}

\newcommand{\efficiencycruise}[0]{             
                      ${\mathcal{J}}_{x, \text{max}}$ ($\text{m/s}^3$)
                        & ${\mathcal{J}}_{y, \text{max}}$ ($\text{m/s}^3$)
                        &${\mathcal{A}}_{x,\text{ma}}$ ($\text{m/s}^2$) 
                        & ${\mathcal{A}}_{y,\text{ma}}$ ($\text{m/s}^2$) }
  \newcommand{\efficiencycruisengsim}[0]{     
                        ${\mathcal{J}}_{x, \text{max}}$ ($\text{m/s}^3$)  
                        & ${\mathcal{J}}_{y, \text{max}}$ ($\text{m/s}^3$)
                      & ${\mathcal{A}}_{x,\text{ma}}$ ($\text{m/s}^2$) 
                       & ${\mathcal{A}}_{y, \text{ma}}$ ($\text{m/s}^2$)           &${\mathcal{\dot{\theta}}}_{\text{ma}}$  ($\text{rad/s}$)
                        }
\newcommand{\efficiencycomputation}[0]{  ${\mathcal{T}}_{\text{opt}}$ ($\text{ms}$)
                        }

\newcommand{\mergeconsistency}[0]{${\mathcal{J}}_{y, \text{ma}}$          ($\text{m/s}^3$)     
                     & ${\mathcal{\dot{\theta}}}_{\text{ma}}$  ($\text{rad/s}$)
                    }
\newcommand{\rom}[1]{(\expandafter{\romannumeral #1\relax})}  
\makeatletter
\pretocmd\@bibitem{\color{black}\csname keycolor#1\endcsname}{}{\fail}
\newcommand\citecolor[1]{\@namedef{keycolor#1}{\color{blue}}}
\makeatother  
\begin{document}
\title{
Safe and Real-Time Consistent Planning for Autonomous Vehicles in Partially Observed Environments via Parallel Consensus Optimization 
} 

\author{Lei Zheng, Rui Yang, Minzhe Zheng, Michael Yu Wang, \textit{Fellow, IEEE,} and Jun Ma, \textit{Senior Member, IEEE}  
  \thanks{This work was supported in part by the Guangdong Basic and Applied Basic Research Foundation under Grant 2025A1515011812; and in part by the Guangdong provincial project under Grant 2023QN10Z006. \textit{(Corresponding author: Jun Ma.)}}
  \thanks{
    Lei Zheng, Rui Yang, and Minzhe Zheng are with the Robotics and Autonomous Systems Thrust, The Hong Kong University of Science and Technology (Guangzhou), Guangzhou 511453, China (email: lzheng135@connect.hkust-gz.edu.cn; ryang253@connect.hkust-gz.edu.cn; mzheng615@connect.hkust-gz.edu.cn).}
    \thanks{Michael Yu Wang is with the School of Engineering, Great Bay University, Dongguan 523808, China (email: mywang@gbu.edu.cn).}
\thanks{Jun Ma is with the Robotics and Autonomous Systems Thrust, The Hong Kong University of Science and Technology (Guangzhou), Guangzhou 511453, China, and also with the Division of Emerging Interdisciplinary Areas, The Hong Kong University of Science and Technology, Hong Kong SAR, China (e-mail: jun.ma@ust.hk).}  
     % \thanks{This work has been submitted to the IEEE for possible publication. Copyright may be transferred without notice, after which this version may no longer be accessible.} 
}   
\markboth{IEEE TRANSACTIONS ON INTELLIGENT TRANSPORTATION SYSTEMS, VOL. 27, NO. 5, May 2026}
{Zheng \MakeLowercase{\textit{et al.}}: SAFE AND REAL-TIME CONSISTENT PLANNING FOR AUTONOMOUS VEHICLES} 

\maketitle  
  
	\begin{abstract}
Ensuring safety and driving consistency is a significant challenge for autonomous vehicles operating in partially observed environments. This work introduces a consistent parallel trajectory optimization (CPTO) approach to enable safe and consistent driving in dense obstacle environments with perception uncertainties.
Utilizing discrete-time barrier function theory, we develop a consensus safety barrier module that ensures reliable safety coverage within the spatiotemporal trajectory space across potential obstacle configurations. Following this, a bi-convex parallel trajectory optimization problem is derived that facilitates decomposition into a series of low-dimensional quadratic programming problems to accelerate computation. By leveraging the consensus alternating direction method of multipliers (ADMM) for parallel optimization, each generated candidate trajectory corresponds to a possible environment configuration while sharing a common consensus trajectory segment. This ensures driving safety and consistency when executing the consensus trajectory segment for the ego vehicle in real time. 
We validate our CPTO framework through extensive comparisons with state-of-the-art baselines across multiple driving tasks in partially observable environments. Our results demonstrate improved safety and consistency using both synthetic and real-world traffic datasets.   
	\end{abstract}

    \begin{IEEEkeywords}
    Autonomous driving, alternating direction method of multipliers, consensus optimization, trajectory planning.
    \end{IEEEkeywords}
      \noindent Video of the experiments: \protect\url{https://youtu.be/YAdtW7J75SY} 
	\section{Introduction}
	\label{sec:introd}
    	\IEEEPARstart{S}{afe} and efficient high-speed navigation for autonomous vehicles in partially observed environments is a formidable challenge \cite{claussmann2020review,schwarting2018planning}. Despite the intensive research on planning and control approaches, the real-time trajectory generation that ensures safety and motion consistency in obstacle-rich environments is still a critical concern~\cite{zheng2024,kousik2021safe}. One of the key underlying factors to this challenge is perception uncertainties, stemming from sensor inaccuracies or unforeseen actions of surrounding vehicles (SVs)~\cite{zhou2024interaction}. These uncertainties can lead to inaccurate obstacle detection, prompting the autonomous high-speed ego vehicle (EV) to perform abrupt maneuvers such as sudden lane changes or decelerations, thereby disrupting driving consistency and compromising task efficiency. Moreover, the nonlinear nature of vehicle dynamics and collision avoidance requirements introduce non-convex constraints in trajectory planning~\cite{ma2022alternating,han2024,huang2023decentralized}. This complexity poses a challenge for real-time replanning over long planning steps, potentially leading to collisions if real-time optimization is not feasible.

   To achieve real-time planning in autonomous driving, a typical approach is to address lateral and longitudinal motions separately, and then combine them to generate a three-dimensional spatiotemporal trajectory for the EV~\cite{sharath2020enhanced, qian2022synchronous, pek2020fail}. Alternatively, researchers have explored decoupling the trajectory's spatiotemporal space into  $(s \times t)$ space for real-time trajectory generation~\cite{chen2022efficient,jian2022multi,chen2024ir,fu2023efficient}.  In~\cite{chen2022efficient}, a learning-based interaction point model is introduced to enhance safety interactions between the EV and SVs. To further account for perception uncertainties, a real-time velocity planner using chance constraints has been proposed \cite{fu2023efficient}. This planner adjusts velocity profiles along a reference path, considering the uncertainty in obstacle occupancy areas. Although these decoupled approaches can facilitate computational efficiency, the quality of the generated trajectory may be affected~\cite{li2024multi}. This issue is particularly evident in dense obstacle environments where effective coordination between spatial and temporal factors is crucial. On the other hand, the model predictive control (MPC) provides an efficient way for spatiotemporal trajectory optimization~\cite{borrelli2017predictive}. To ensure safety for autonomous vehicles, the control barrier functions (CBFs)~\cite{ames2017control,zhao2023safety,zheng2024incremental} have been leveraged to construct safety constraints in nonlinear MPC frameworks~\cite{zeng2021safety,son2019safety}. 
    Despite their effectiveness, these methods face computational challenges, primarily due to the inversion of the Hessian matrix during optimization. To tackle this issue, an efficient FITS approach in quadratic programming (QP) form has been proposed~\cite{vahs2024forward}. This method leverages CBF to ensure safety in the spatiotemporal trajectory space. Additionally, multiple shooting techniques have been employed to facilitate optimization within the MPC framework under dense traffic~\cite{zheng2024}. However, {these approaches typically generate a single locally optimal trajectory based on fixed predictions of the intentions of SVs, without considering potentially inaccurate perceptions of the environment. This limitation can compromise driving stability and pose a safety threat to the EV in partially observed environments.}
  
    On the other hand, parallel trajectory generation approaches have been proposed to optimize multiple candidate trajectories in autonomous driving~\cite{li2024multi, batkovic2021robust, chen2022interactive}. Considering multi-modal behaviors of road users, a robust scenario MPC is developed to optimize over a tree of trajectories with time-varying feedback policies \cite{batkovic2021robust}. To further account for the interaction between the EV and SVs, a risk-aware branch MPC is introduced with a scenario tree considering a finite set of potential motions of the uncontrolled agent in highway driving \cite{chen2022interactive}. This approach solves a feedback policy in the form of a trajectory tree with multiple branch points to account for the future motion of SVs. Alternatively, partially observable Markov decision process approaches~\cite{li2023pomdp,tang2022integrated,indelman2015planning} are employed to deal with the uncertain behaviors of SVs over a similar tree structure in autonomous driving. {While these approaches show promise in handling environmental uncertainties, they are computationally intensive as the problem size increases \cite{li2023marc}. }
 
     To streamline the optimization process in autonomous driving, researchers have employed parallel computation techniques to accelerate trajectory optimization~\cite{zheng2023real, adajania2022multi, liu2024improved, zheng2024barrier}. In~\cite{zheng2023real}, multi-threading techniques have been utilized to optimize each trajectory on distinct cores under congested traffic scenarios. For highway applications, the Batch-MPC framework~\cite{adajania2022multi} employs alternating minimization~\cite{tseng1991applications} to decompose complex multi-trajectory optimization problems into manageable subproblems. To further address the feasibility of optimization with inequality constraints, the over-relaxed alternating direction method of multipliers (ADMM)~\cite{ghadimi2015optimal} has been explored for parallel trajectory optimization in road construction scenarios~\cite{zheng2024barrier}. Although these methods facilitate high computational efficiency by optimizing each trajectory on separate threads or as distinct low-dimensional optimization problems, they lack coordination among the different candidate trajectories. This could affect driving safety and consistency, especially when switching between locally optimal trajectories under perception uncertainties. To enhance driving consistency under perception uncertainties, researchers have extensively implemented contingency planning~\cite{li2023marc, alsterda2019contingency, packer2023anyone} in autonomous driving with partial observability. These approaches generate a set of possible motions in a scenario-tree structure, sharing a common trajectory trunk during execution. To achieve fast distributed optimization, {the Control-Tree approach based on distributed ADMM has been developed for autonomous driving in partially observable environments~\cite{phiquepal2021control}. While multi-threading improves the efficiency of ADMM subproblem processing, overall computation remains intensive, particularly in scenarios with dense interactions.}
  
    In this paper, we present a \textbf{C}onsistent \textbf{P}arallel \textbf{T}rajectory \textbf{O}ptimization (CPTO) framework to achieve real-time, consistent, and safe trajectory planning for autonomous driving in partially observed environments. Our approach introduces a spatiotemporal consensus safety barrier, ensuring that each candidate trajectory aligns with a potential obstacle configuration. {Different from \cite{zheng2023real, adajania2022multi, zheng2024barrier}, CPTO ensures that all optimized candidate trajectories share the same consensus topology segment for coordination before diverging. This facilitates driving consistency and strikes a balance between safety and task efficiency.  
     Furthermore, compared to branch-based MPC and contingency planners~\cite{phiquepal2021control,chen2022interactive,batkovic2021robust,li2023marc}, CPTO directly transforms non-convex constraints through bi-convex transcription and resolves them via a consensus-ADMM scheme. This decomposition generates a series of convex subproblems that enable fast optimization while preserving solution fidelity. }

 The main contributions of this paper are summarized as follows:
\begin{itemize} 
    \item   We introduce a consensus safety barrier module to ensure reliable safety coverage in trajectory space under perception uncertainties. This module allows each generated trajectory to share a common consensus segment while accounting for different scenarios, ensuring driving safety and consistency in dense traffic. By utilizing discrete-time barrier function theory, we guarantee forward invariance of the generated trajectory. 
    \item We exploit the bi-convex nature of the constraints and use parallel consensus ADMM iterations to transform the non-convex NLP planning problem into a series of low-dimensional QP problems. This strategy ensures each generated feasible trajectory adheres to the same consensus segment while enabling large-scale optimization in real time.
    \item We validate the effectiveness of our algorithm by comparing it with other state-of-the-art baselines in partially observable environments across various driving tasks. Additionally, we investigate the influence of consensus steps and provide a detailed computation time analysis.
\end{itemize}

    The rest of this paper is structured as follows.  Section~\ref{sec:problem} provides necessary preliminaries for parallel trajectory optimization and outlines the problem statement. Section~\ref{sec:safety} presents a spatiotemporal control barrier module for safe navigation, accompanied by rigorous proof. Section~\ref{subsec:CPTO} details the general optimization procedure for trajectory generation using parallel consensus optimization. The effectiveness of our algorithm is demonstrated in partially observed environments with a detailed analysis of performance trade-offs across different consensus steps in Section~\ref{sec:exp}. Finally, Section~\ref{sec:con} concludes the paper.

  \section{Preliminaries and Problem Statement} 
    \label{sec:problem}  

\subsection{{Notation}}  
\label{subsec:notation}  
{This section introduces the notation used throughout the study. The set of real numbers is denoted by \( \mathbb{R} \), and \( \mathbb{R}^n \) represents the \( n \)-dimensional Euclidean space. The \( p \)-norm of a vector or matrix is denoted by \( \|\cdot\|_p \),  and \( \|\cdot\| \) specifically refers to the Euclidean norm for vectors and the spectral norm for matrices.  }

{ Given a vector \( \mathbf{z} \in \mathbb{R}^n \),
the inequality \( \mathbf{z} \succeq  \mathbf{0} \) indicates that each component of \( \mathbf{z} \) is non-negative. Likewise, \( \mathbf{z} \preceq \mathbf{0} \) means each component of \( \mathbf{z} \) is less than or equal to zero, \( \mathbf{z} \succ \mathbf{0} \) means each component is greater than zero, and \( \mathbf{z} \prec \mathbf{0} \) means each component is less than zero. These notations are used to express element-wise inequalities, ensuring that each component of the vector or matrix is compared individually. The indicator function \( \mathcal{I}_{+}(\cdot) \) is defined as:  
\[
\mathcal{I}_{+}(\mathbf{z}) = \begin{cases}
    \mathbf{0}, & \text{if } \mathbf{z} \succeq \mathbf{0}, \\
    +\infty, & \text{otherwise}.
\end{cases}
\]  
This function enforces non-negativity constraints in the optimization process.  }

{In matrix operations, \( \otimes \) denotes the vertical stacking operator, which concatenates matrices or vectors vertically. For example, if \( \mathbf{A} \in \mathbb{R}^{m \times n} \) and \( \mathbf{B} \in \mathbb{R}^{p \times n} \), then \( \mathbf{A} \otimes \mathbf{B} \in \mathbb{R}^{(m+p) \times n} \). \( \cdot \) denotes the Hadamard product, which performs element-wise multiplication. }

{The set of integers from \(0\) to \(N\) is denoted by \( \mathcal{I}^{N}_0 = \{0, 1, \dots, N\} \). The four-quadrant arctangent function \( \arctan(\cdot) \) computes the angle between the EV and obstacles. The function \( \max(\cdot) \) ensures that safety margins \( d \) do not fall below a specified threshold (e.g., \( d \geq 1 \)) by performing element-wise maximum value clipping. The set intersection operator \( \cap \) defines the consensus safe set \( \mathcal{C}_{\text{cons}} \).} {Table~\ref{tab:symbol} summarizes the nomenclature used in this paper.   }

\begin{table}[t]  
\centering  
\scriptsize  
\caption{{Nomenclature} } 
\label{tab:symbol}  
\begin{tabular}{@{}ll@{}}  
\toprule  
Symbol & Description \\  
\midrule  
\( \mathcal{X} \) & State space of the EV \\  
\( \mathbf{C}_x, \mathbf{C}_y, \mathbf{C}_\theta \) & Control point variables for Bézier trajectories \\  
\( \mathbf{L}_x, \mathbf{L}_y \) & Semi-major and semi-minor axes of the safety ellipse \\  
\( \mathbf{Z}_x, \mathbf{Z}_y \) & Slack variables for inequality constraints \\  
\( \mathbf{Y}_x, \mathbf{Y}_y, \mathbf{Y}_\theta \) & Consensus variables for shared trajectory segments \\  
\( \mathbf{D} \) & Safety margin scaling factors  \\  
\( \boldsymbol{\omega} \) & Relative angles between the EV and obstacles \\  
\( \mathcal{O} \) & Set of obstacle configurations \\  
\( \mathcal{C}_{\text{cons}} \) & Consensus safe set (forward invariant) \\  
\( \mathbf{Q}_P \) & Weighting matrix for trajectory smoothness \\  
\( \boldsymbol{\alpha} \) & Barrier coefficient for safety adjustments \\  
\( \boldsymbol{\lambda} \) & Dual variables for ADMM optimization (e.g., \( \boldsymbol{\lambda}_{\text{cons}}, \boldsymbol{\lambda}_{\text{obs}} \)) \\  
\( \rho \) & Penalty parameters for ADMM (e.g., \( \rho_{\text{cons}}, \rho_{\text{obs}} \)) \\  
\( N \) & Number of planning steps in a planning horizon \\  \( N_c \) & Number of candidate trajectories \\  
\( N_s \) & Consensus steps (shared trajectory segment) \\  
\( \mathcal{I}^{N}_0 \) & Set of integers \( \{0, 1, \dots, N\} \) \\  
\bottomrule  
\end{tabular}  
\end{table}

    \subsection{Motion Model and State Augmentation}   
   In this study, we adopt Dubin’s car model as the motion model for the red EV, which uses yaw rate $\dot{\theta}$ and acceleration as control inputs~\cite{chen2024interactive}.  
    To facilitate smooth trajectory optimization, we augment the state vector of the EV to include the control inputs and their derivatives. The augmented state vector is defined as follows:  
\begin{equation} \vspace{-0mm}
   \mathbf{x} = [p_x\quad p_y \quad \theta \quad \dot{\theta} \quad v \quad a_x\quad a_y\quad j_x\quad j_y ]^T \in \mathcal{X}, 
     \label{eq:state_vector}
\end{equation} 
where $p_x$ and $p_y$ denote the longitudinal and lateral positions of the EV in the global coordinate system, respectively; $\theta$ denotes the heading angle of the EV; \(v\) denotes the velocity of the EV in the global coordinate system; \(a_x\) and \(a_y\) represent the longitudinal and lateral accelerations in the global coordinate system, respectively; and \(j_x\) and \(j_y\) represent the longitudinal and lateral jerks in the global coordinate system, respectively. 

Following \cite{vahs2024forward}, we describe the state evolution along the \(j\)-th augmented state trajectory \(\mathcal{T}_j^\mathcal{K}\) over the closed interval \(\mathcal{K} = [t_0, t_0+T]\) as:
\begin{equation}
\mathcal{T}_j^\mathcal{K} := \{  \mathbf{x}(t_k) \in \mathcal{X} \mid t_k \in \mathcal{K}\} \subseteq \mathcal{X}^\mathcal{K} \subseteq L^2(\mathcal{K}, \mathcal{X}),    \label{eq:trajectory}
\end{equation}  
where \(\mathcal{X}^\mathcal{K}\) is the set of possible trajectories that are square integrable \((L^2)\) over the interval \(\mathcal{K}\).  Here, \(t_0\) denotes the initial time, and \(t_k = t_0 + k\delta t\) represents the current time instant, with \(k\) being the current time step.  The discrete-time interval is given by \(\delta t = T/N\), where \(N\) is the total number of planning steps, and \(T\) represents the time interval, typically representing the planning horizon. For clarity, $ \mathbf{x}_{t_0,k} $ is used to denote $ \mathbf{x}(t_k) $ from the initial time $t_0$. 
  \begin{figure}
		\centering
		\includegraphics[width=8.5cm]{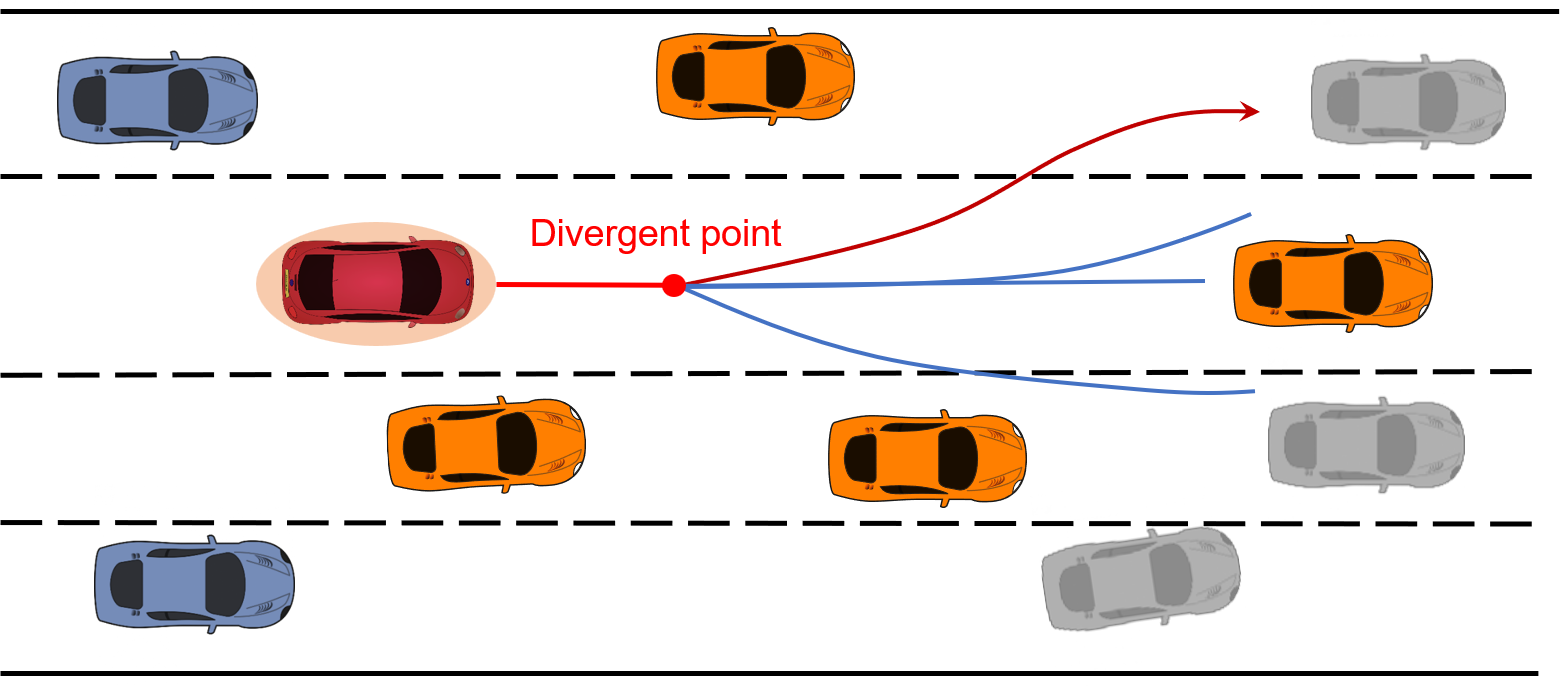}\vspace{-0mm}
		\caption{Illustration of the motion of the EV (in red) in a dense traffic scenario under perception uncertainties. The orange and grey vehicles represent observed and uncertain vehicles considered in the current planning framework, respectively. The blue vehicles denote unconsidered vehicles in the current planning framework. The red curve indicates the selected trajectory for the EV to execute at the current time instant, and the blue curves represent other generated trajectories. All generated trajectories share a common segment and diverge at a specific divergent point to account for different scenarios, ensuring safe and consistent maneuvering under perception uncertainties.} \vspace{-0mm}
  \label{fig:Problem_statement}
	\end{figure} 
\subsection{ Trajectory State Parameterization}
\label{subsec:Trajectory_para} 
 In this study, we utilize three-dimensional  B\'ezier curves to represent the trajectories of the augmented state \eqref{eq:state_vector}. These curves capture the evolution of the EV's state in terms of the longitudinal direction, lateral direction, and orientation. For the $j$-th trajectory, a  B\'ezier curve of degree \(n\)~\cite{farouki2012bernstein,tong2023} is defined by \(n+1\) control points over the interval \(\nu \in [0, 1]\) as follows: 
\begin{equation}\vspace{-0mm}
    \mathbf{C}^{(j)}(\nu) = \sum_{i=0}^{n} \mathbf{p}^{(j)}_{i}B_{i,n}(\nu), \quad j \in \mathcal{I}_0^{N_c-1}, 
    \notag \vspace{-0mm}
\end{equation}  
where \(\mathbf{p}^{(j)}_{i} = [c^{(j)}_{x,i} \quad c^{(j)}_{y,i} \quad c^{(j)}_{\theta,i}]^T \in \mathbb{R}^3\) represents the control points for the \(j\)-th trajectory; \(B_{i,n}(\nu) = \binom{n}{i} \nu^i (1 - \nu)^{n - i}\) is the Bernstein polynomial basis, where the parameter \(\nu\) is defined as \(\nu = \frac{t_k - t_0}{T} \in [0, 1]\). The resulting trajectory sequences can be expressed as:
\[
\left\{ \mathbf{C}^{(j)}_k \right\}_{k=0}^{N-1} = \mathbf{P}^T_{j} \mathbf{W}_B, \quad j \in \mathcal{I}_0^{N_c-1},
\] 
where \(\mathbf{C}^{(j)}_k = [p^{(j)}_{x,k} \quad p^{(j)}_{y,k} \quad \theta^{(j)}_{k}]^T\) denotes the longitudinal position, lateral position, and heading angle for the \(j\)-th trajectory at time step \(k\). The matrix of control points to be optimized is \(\mathbf{P}_{j} = [\mathbf{p}^{(j)}_0 \quad \mathbf{p}^{(j)}_1 \quad \dots \quad \mathbf{p}^{(j)}_n]^T \in \mathbb{R}^{(n+1) \times 3}\); The constant basis matrix \(\mathbf{W}_B = [\mathbf{B}_{0} \quad \mathbf{B}_{1} \quad \dots \quad \mathbf{B}_{n}]^T \in \mathbb{R}^{(n+1) \times N}\), where \(\mathbf{B}_{i} = [B_{i,n}(\delta t) \quad B_{i,n}(2\delta t) \quad \dots \quad B_{i,n}(N\delta t)]^T \in \mathbb{R}^N\), and $\mathbf{B}_{i,k} =  B_{i,n}(k\delta t)$  for all $k \in \mathcal{I}_1^{N}$.  
Note that the hodograph property of  B\'ezier curves facilitates constraining high-order
derivatives of the  B\'ezier curve, inherently enforcing dynamical constraints on the EV's motion. 
 \subsection{ Problem Statement} 
 \label{subsection:problem_statement}
This study considers multi-lane dense and cluttered driving scenarios in partially observed environments, as depicted in Fig.~\ref{fig:Problem_statement}. In such scenarios, obstacle detection is prone to inaccuracies, leading to false positives, wherein obstacles suddenly appear or disappear. This necessitates a real-time motion planning strategy that allows the EV to navigate safely among multiple obstacles without compromising its task performance.  To render this problem tractable, we adopt the following perception assumption:
 \begin{assumption}(\textbf{Sensing \ Capabilities})
    % ~\cite{shalev2017formal})
    \label{assumption: observation}
In autonomous driving, the EV can fully observe an obstacle once it comes within a certain threshold distance of the EV.  
    \end{assumption} 
 
We define the objective function as the sum of the weighted squared \(L_2\) norms of the control point matrices, as follows: 
\begin{alignat}{2}
    \small
    \mathcal{L}  = \sum_{j=0}^{N_c-1} q_j||\mathbf{P}_{j} ||^2_2,
    \label{eq:obj_func} \vspace{-0mm}
\end{alignat} 
where \(q_j\) is a constant positive weight, $N_c$ denotes the total number of generated trajectories. The motion planning problem is then formulated as follows:  
\begin{subequations}
    \label{problem}\begin{align}
    \displaystyle\operatorname*{minimize}~~~
    & \mathcal{L}\label{eq:problem}\\
    \quad\text{subject to}\quad
    &\mathbf{x}^{(j)}_{t_0,0}=\mathbf{x}(t_0), \quad \mathbf{x}^{(j)}_{t_0,N} \in \mathcal{X}^{(j)}_f, \label{eq:problem_1}\\
    & \mathbf{x}^{(j)}_{t_0,k+1}  = f^{EV}(\mathbf{x}^{(j)}_{t_0,k}),  \label{eq:problem_2}\\ 
    & \mathbf{O}^{(i)}_{k+1}= \zeta (\mathbf{O}^{(i)}_k) +  w^{(i)}_k, \quad i\in\mathcal{I}_0^{m-1},\label{eq:problem_3}\\ 
    & \mathbb{X}^{(j)}_k \cap \mathbb{O}^{(i)}_k = \emptyset, \quad i\in\mathcal{I}_0^{m-1},\label{eq:problem_4}\\
    & \varphi( \mathbf{x}^{(j)}_{t_0,l}) = \varphi(\Tilde{\mathbf{x}}_{t_0,l}), \quad \forall l\in\mathcal{I}_0^{N_s-1}, \label{eq:problem_5}\\ 
    & \mathbf{x}^{(j)}_{t_0,k} \in  \mathcal{X}, \label{eq:problem_6}\\ 
    & \forall k\in\mathcal{I}_0^{N-1},  \quad j\in\mathcal{I}_0^{N_c-1}. \nonumber \vspace{-0mm}
\end{align}
\end{subequations} 
Here, \(\mathbf{x}(t_0)\) denotes the initial state of the EV, while \(\mathcal{X}^{(j)}_f\) denotes the terminal set for the $j$-th trajectory. The function \(f^{EV}(\cdot)\) captures the EV's kinematic constraints, and \(\zeta(\cdot)\) describes the state evolution of the \(i\)-th obstacle. 
The kinematic constraint \eqref{eq:problem_2} of the EV is governed by the hodograph property of the  B\'ezier curve and the following nonholonomic constraints, as outlined in~\cite{zheng2024barrier,adajania2022multi}:
    \begin{eqnarray}
    \left\{
        \begin{aligned}
          \dot{p}^{(j)}_{x,k} -& v^{(j)}_k \cos (\theta^{(j)}_k) = 0, \forall  k\in\mathcal{I}_0^{N-1}, j\in\mathcal{I}_0^{N_c-1},\\
           \dot{p}^{(j)}_{y,k}- &v^{(j)}_k\sin (\theta^{(j)}_k) = 0,  \forall  k\in\mathcal{I}_0^{N-1}, j\in\mathcal{I}_0^{N_c-1},
        \end{aligned}
    \right.
    \label{eq:nonholonomic_cons}
    \end{eqnarray}   
where ${p}^{(j)}_{x,k}$ and ${p}^{(j)}_{y,k}$ denote the longitudinal and lateral state of the EV at time step $k$ for generating the $j$-th trajectory. The same convention applies to other states of the EV in \eqref{eq:state_vector}. 

 The  position and velocity state of the $i$-th obstacle at time step $k$ is denoted by $\mathbf{O}^{(i)}_k =  [o_{x,k}^{(i)} \quad o_{y,k}^{(i)} \quad o_{v_x,k}^{(i)} \quad o_{v_y,k}^{(i)}]^T $. The term \(w^{(i)}_k\) in \eqref{eq:problem_3} accounts for the observation noise associated with the \(i\)-th obstacle, which is generally bounded and decreases as the EV approaches the obstacle.  

The condition in \eqref{eq:problem_4} ensures collision avoidance during the planning process, where the EV and the $i$-th SV are represented as an ellipse-shaped convex compact set $\mathbb{X}^{(j)}$  and $\mathbb{O}_i $, respectively; $m$ denotes the number of obstacles considered in the current planning framework, which may vary due to uncertainties in obstacle existence;  \(\mathbf{x}_{k,\min}\) and \(\mathbf{x}_{k,\max}\) denote the minimum and maximum limits for the augmented state constraints, ensuring adherence to the state and actuator limitations of the EV. 

The constraint in \eqref{eq:problem_5} is the consensus constraint, which enforces each generated trajectory to share a consensus segment before diverging to enhance driving consistency. Here, $N_s < N$ denotes the consensus step, and function $\varphi (\cdot)$ extracts the desired variable from the global consensus variable $\Tilde{\mathbf{x}}_{t_0,l}$. It extracts the position, velocity, acceleration, and orientation states of the EV to enforce consensus constraints in this study, ensuring uniformity in these aspects across all trajectories.  
  
 We employ the evaluation strategy outlined in Section V of~\cite{zheng2024barrier} to prioritize driving safety, passenger comfort, and task accuracy. This enables the EV to select the optimal trajectory among all candidate trajectories to execute. Note that the chosen trajectory will inherently consider the objectives of all candidate trajectories, as these trajectories share a common segment before diverging during the receding horizon planning.
  
\section{Consensus Spatiotemporal Safety Barrier}
\label{sec:safety}
In this section, we develop a consensus-based spatiotemporal safety barrier to ensure safe and efficient interactions for the EV in partially observed environments.

\subsection{Polar Representation of Safety Constraints} 
Referring to the approach outlined in~\cite{rastgar2020novel,zheng2024barrier}, we convert the Cartesian coordinates of the EV and SVs into polar coordinates. This transformation allows us to represent the Euclidean distance and relative angle between the EV and SVs more effectively. 
To ensure safety, we implement the following polar-based safety constraints:  
\begin{eqnarray}
\left\{
    \begin{aligned}
     p_{x,k} &= o_{x,k}^{(i)} + l_{x,k}^{(i)} d_k^{(i)} \cos(\omega_k^{(i)}), \\
     p_{y,k} &= o_{y,k}^{(i)} + l_{y,k}^{(i)} d_k^{(i)} \sin(\omega_k^{(i)}), \\
     d^{(i)}_k & \geq 1, \quad \forall i\in\mathcal{I}_0^{m-1},
    \end{aligned}
\right.
\label{eq:polar_safety1}
\end{eqnarray}  
where \(l_{x,k}^{(i)}\) and \(l_{y,k}^{(i)}\) represent the semi-major and semi-minor axes of the safety ellipse, respectively; \(d_k^{(i)}\) is the scaling factor for the \(i\)-th obstacle. We can further derive the closed-form value of the angle \(\omega_k^{(i)}\) between the EV and the \(i\)-th SV as:  
\begin{equation} \vspace{-0mm}
    \omega^{(i)}_k = \arctan\left(l_{x,k}^{(i)}( p_{y,k}  - o_{y,k}^{(i)}), \, l_{y,k}^{(i)}( p_{x,k}  - o_{x,k}^{(i)})\right).
    \label{eq:angle_value}
 \vspace{-0mm} \end{equation}

Note that the polar-based safety constraints can inherently account for sensor noise in both longitudinal and lateral directions. By dynamically adjusting the semi-major and semi-minor axes of the safety ellipse across different steps $k$ in the planning horizon, the bounded noise \(w^{(i)}_k\) in \eqref{eq:problem_3} can be accommodated. The safety ellipses are designed to shrink over time as the EV approaches an obstacle, employing a linearly decreasing function. This approach ensures that the safety margin decreases from an initial maximum constant value $l_{x,\max}$ and $l_{y,\max}$ to a defined minimum safety threshold $l_{x,\min}$ and $l_{y,\min}$, enabling the EV to navigate safely without exhibiting overly conservative behaviors. Additionally, the condition \(d^{(i)}_k \geq 1\) guarantees that this safety margin is consistently maintained, effectively preventing collisions based on Assumption 1.

\subsection{Consensus Safety Barrier in Trajectory Space}   
\label{subsec:consenus_safety}
To facilitate safe and efficient navigation for the EV, it is crucial to address the existence uncertainty of obstacles, especially in dense traffic where obstacles may not be fully observed. This requires considering the presence of obstacles under perception uncertainties. Additionally, it is important to balance safety and task performance while avoiding overly conservative collision avoidance behaviors to enable the EV to accomplish its tasks effectively.  
\subsubsection{Safety Barrier} 
We assume that it is sufficient to consider constraints at the discrete time steps along the planned trajectory $\mathcal{T}_j^\mathcal{K}$, as discussed in~\cite{vahs2024forward}. To provide formal safety guarantees for the EV in the $j$-th trajectory space, we define the following safe set $ \mathcal{C}^{(j)} $: 
\begin{subequations} \label{eq:safe_set}
\begin{align}
\mathcal{C}^{(j)} & := \{\mathbf{s}^{(j)}_{t_k,[0,N]} \in \mathcal{T}_j^\mathcal{K} \mid \mathbf{h}(\mathbf{s}^{(j)}_{t_k,[0,N]}, o^{(j)}) \succeq  \mathbf{0}\}, \label{eq:safe} \\
Out(\mathcal{C}^{(j)}) & := \{\mathbf{s}^{(j)}_{t_k,[0,N]} \in \mathcal{T}_j^\mathcal{K} \mid \mathbf{h}(\mathbf{s}^{(j)}_{t_k,[0,N]}, o^{(j)}) \prec \mathbf{0}\}, \label{eq:unsafe}
\end{align}
\end{subequations}
where  \(o^{(j)} \in \mathcal{O}\) represents a potential arrangement of obstacle configurations for the \(j\)-th trajectory, with \(j \in \mathcal{I}_0^{N_c-1}\).  The variable $\mathbf{s}^{(j)}_{t_k,[0,N]}$ represents the generated discrete trajectory sequence with $N$ steps at the current time $t_k$, defined as follows:
\begin{equation}
   \mathbf{s}^{(j)}_{t_k,[0,N]}:= [\mathbf{x}^{(j)}_{t_k,0}\quad \mathbf{x}^{(j)}_{t_k,1}\quad \cdots \quad \mathbf{x}^{(j)}_{t_k,N}]^T \in \mathcal{S} \subseteq  \mathcal{T}_j^\mathcal{K};\notag
\end{equation}
\begin{equation}\mathbf{h}(\mathbf{s}^{(j)}_{t_k,[0,N]}, o^{(j)}) = [h(\mathbf{x}^{(j)}_{t_k,0},o^{(j)})  \quad \cdots \quad h(\mathbf{x}^{(j)}_{t_k,N},o^{(j)})]^T, \notag
\end{equation}
where each element $h:\mathcal{X}  \times \mathcal{O} \to \mathbb{R} $ is a \(C^1\) function that accounts for safety constraints in~\eqref{eq:problem_4} at each time step. Thus, we can obtain the following safe set for all trajectories:
\begin{equation}
    \mathcal{C}_h = \bigcap_{j=0}^{N_c-1}  \mathcal{C}^{(j)}.
    \label{eq:safe_sets1}\vspace{-2mm}
\end{equation}

\begin{definition}(\cite{zeng2021enhancing})
\label{def:discrete_cbf}
 A $C^1$ function \( h: \mathcal{X} \times \mathcal{O} \to \mathbb{R} \) is said to be a discrete-time  barrier function (BF) for the set $\mathcal{C}^{(j)} \subset  \mathcal{T}_j^\mathcal{K}$ in \eqref{eq:safe_set}, if it satisfies the following condition:
 \begin{equation} \vspace{-0mm}\label{eq:discrete_cbf}
 \Delta   h(\mathbf{x}^{(j)}_{t_k},o^{(j)})  > -\gamma_k( h(
\mathbf{x}^{(j)}_{t_k},o^{(j)} )  ),\quad\forall j \in \mathcal{I}_0^{N_c-1},
 \vspace{-0mm} \end{equation}
where \(\Delta h(\mathbf{x}^{(j)}_{t_k},o^{(j)})  :=  h(\mathbf{x}^{(j)}_{t_k+ \delta t},o^{(j)}) -    h(\mathbf{x}^{(j)}_{t_k},o^{(j)})\); $\gamma_k$ is an extended $\mathcal{K}$ function, i.e., continuous and strictly increasing with  \(\gamma_k(0) = 0\), and \( \gamma_k (h)  = \alpha_k  h, \alpha_k \in (0, 1), \forall h \neq 0\).
\end{definition}   
\begin{remark} \label{remark:1}
   Large values of \(\alpha_k\) contribute to a more aggressive adjustment of the trajectory sequence, and vice versa~\cite{zeng2021safety}. As perception uncertainties decrease when obstacles approach the EV, we gradually increase barrier coefficients \(\alpha_k\) to facilitate more stable trajectory adjustments in the near future and avoid overly conservative behaviors in the later stages. This strategy strikes a balance between safety and task performance.
\end{remark}  
% { \begin{remark} \label{remark:2}  
% Given \(\alpha_k\), the BF \(h: \mathcal{X} \times \mathcal{O} \to \mathbb{R} \) acts as an adaptive safety margin:  
% \begin{itemize}  
%   \item When \(h\) is large (system far from the safety boundary), \(\gamma_k(h)\) is large, allowing flexibility in trajectory planning.  
%   \item When \(h\) is small (near the safety boundary), \(\gamma_k(h)\) shrinks. This prevents rapid decreases in \(h\).  
% \end{itemize}  
% \end{remark}   }

{
\begin{remark} \label{remark:2}
Given \(\alpha_k\), the barrier function \(h: \mathcal{X} \times \mathcal{O} \to \mathbb{R}\) generates an adaptive safety margin via its scalar output:
\begin{itemize}
  \item A large safety margin (far from the safety boundary) produces a large \(\gamma_k(h)\), allowing flexible trajectory planning.
  \item A small safety margin (near the boundary) reduces \(\gamma_k(h)\), preventing rapid margin decreases to preserve safety.
\end{itemize}
\end{remark}}

Referring to~\cite{zheng2024barrier}, we leverage the definition of BF and the polar-based safety constraints in~\eqref{eq:polar_safety1} to transform the original safety constraints into the following spatiotemporal safety constraints in trajectory space $\mathcal{T}_j^\mathcal{K}$:
\begin{eqnarray}
\left\{
    \begin{aligned}
     & \mathbf{p}_{x}  = \mathbf{o}_{x}^{(j)} + \mathbf{l}^{(j)}_x \cdot  \mathbf{d}^{(j)}   \cdot  \cos ( \bm{\omega}^{(j)}), \\
    & \mathbf{ p}_{y} = \mathbf{o}^{(j)}_{y} +\mathbf{l}^{(j)}_y \cdot  \mathbf{d}^{(j)}   \cdot 
 \sin ( \bm{\omega}^{(j)}),  \\
      & \Delta  \mathbf{h}(\mathbf{s}^{(j)}_{t_k,[0,N]},o^{(j)})  \succ -  \bm{\alpha}^{[0,N]} \cdot \mathbf{h}(\mathbf{s}^{(j)}_{t_k+ \delta t,   [0,N]},o^{(j)}),  \\ 
    \end{aligned}
\right.
\label{eq:polar_safety2} \vspace{-0mm}
\end{eqnarray}  
where \begin{equation}
    \label{eq:polar_safety3}
    \mathbf{h}(\mathbf{s}^{(j)}_{t_k,[0,N]}, o^{(j)}) = \mathbf{d}^{(j)} - 1,\quad \forall j \in \mathcal{I}_0^{N_c-1}.\vspace{-0mm}
\end{equation} 
Here, the variable \(\mathbf{d}^{(j)}\) represents a matrix that vertically stacks the scaling factor vector \(d_k^{(i)}\) for all \(  i\in \mathcal{I}_0^{M_j-1}\) along the planning step \(N\); \(M_j\) denotes the number of considered obstacles for the $j$-th trajectory or a given configuration \(o^{(j)} \in \mathcal{O}\). The same convention applies to other variables in \eqref{eq:polar_safety1}. 
The variable \(\bm{\alpha}^{[0, N]} \in \mathbb{R}^{N}\) represents the barrier coefficient for each element \(\alpha_k\) in the \(k\)-th step of a planned horizon with $N$ steps, which increases gradually along the steps \(k\), as discussed in Remark~\ref{remark:1}.  

\subsubsection{Consensus Safety Barrier} 
\label{subsubsec:consensus_barrier}
In partially observed environments, accounting for the presence of uncertain obstacles is essential for generating trajectories. Each trajectory must be adaptable to various potential scenarios.  We define a set of possible obstacle configurations \(\mathcal{O}\), where each configuration \(o^{(j)} \in \mathcal{O},\ \forall j  \in \mathcal{I}_0^{N_c-1}\) represents a potential arrangement of obstacles for the $j$-th trajectory. 

In a receding horizon planning framework, the generated trajectories are structured as follows: 
\begin{equation}
     [ \underbrace{ \mathbf{x}^{(j)}_{t_k,0}}_{\text{initial state}} \quad \underbrace{\varphi(\mathbf{x}^{(j)}_{t_k,[1:N_s]})}_{\text{consensus segment}} \quad \underbrace{\mathbf{x}^{(j)}_{t_k,[N_s+1:N-1]}}_{\text{divergent segment}}].
\end{equation} 
We can guarantee the safety of the EV across all possible obstacle configurations \(o^{(j)} \in \mathcal{O}\), if the consensus segment of the trajectory lies within the safe set \(\mathcal{C}_h\) \eqref{eq:safe_sets1}. To achieve this goal, we formulate the following consensus spatiotemporal BF condition in the trajectory space $\mathcal{S}$:
    
\begin{theorem}
\label{theorem:1}
For all possible obstacle configurations \(o^{(j)} \in \mathcal{O},\ \forall j \in \mathcal{I}_0^{N_c-1}\), if each generated trajectory satisfies the spatiotemporal safety constraints in \eqref{eq:polar_safety2} and the following consensus condition:  
\begin{subequations} \label{eq:consensus_condition}
\begin{align}
& \varphi(\mathbf{x}^{(0)}_{t_k,[1:N_s]})   = \varphi(\mathbf{x}^{(1)}_{t_k,[1:N_s]}) = \dots = \varphi(\mathbf{x}^{(N_c-1)}_{t_k,[1:N_s]}), \label{eq:consensus_condition1} \\
 &h(\varphi(\mathbf{x}^{(0)}_{t_k,[1:N_s]}), o^{(j)} ) = h(\mathbf{x}^{(0)}_{t_k,[1:N_s]},  o^{(j)}),\label{eq:consensus_condition2}
\end{align}
\end{subequations} 
then the following consensus set \( \mathcal{C}_{\text{cons}} \) is \textit{forward invariant}:
\begin{eqnarray}
\left\{
    \begin{aligned}
        \mathcal{C}_{\text{cons}} &= \bigcap_{j=0}^{N_c-1} \mathcal{C}_{\text{cons}}^{(j)} \subseteq \mathcal{C}_{h}, \\
        \mathcal{C}_{\text{cons}}^{(j)} &:= \{\mathbf{s}^{(j)}_{t_0,[0,N_s]} \in \mathcal{T}_j^\mathcal{K} \mid \mathbf{h}(\mathbf{s}^{(j)}_{t_0,[0,N_s]}, o^{(j)}) \succeq \mathbf{0}\}.
    \end{aligned}
\right.
\label{eq:consensus_set}
\end{eqnarray}
\end{theorem}

\begin{IEEEproof} 
Given an initial safe trajectory sequence $\mathbf{s}^{(j)}_{t_0,[0,N_s]} \in \mathcal{C}_{h}$ with $\mathbf{h}(\mathbf{s}^{(j)}_{t_0,[0,N_s]},o^{(j)}) \succ \mathbf{0}$, we can  derive the following inequality based on the spatiotemporal safety constraints in \eqref{eq:polar_safety2}:
\begin{equation}
   \Delta \mathbf{h}(\mathbf{s}^{(j)}_{t_0+ \delta t,[0,N_s]}, o^{(j)})\succ    -  \bm{\alpha}^{[0,N_s]} \cdot \mathbf{h}(\mathbf{s}^{(j)}_{t_0,[0,N_s]},o^{(j)}), \notag
\end{equation} where $\bm{\alpha}^{[0,Ns]} = [\alpha_{0}\quad \alpha_{1}\quad \cdots \quad \alpha_{N_s}]^T$, with each element $\alpha_k \in (0, 1), \forall h \neq 0$. 

As a result, the following inequality holds: 
\begin{align} \vspace{-0mm}
\begin{aligned}
  \mathbf{h}(\mathbf{s}^{(j)}_{t_0+ \delta t,[0,N_s]},o^{(j)}) \succ  (\mathbf{1}-\bm{\alpha}^{[0,N_s]})  \cdot \mathbf{h}(\mathbf{s}^{(j)}_{t_0,[0,N_s]},o^{(j)})\succ\mathbf{0}.
\end{aligned} \notag
\vspace{-0mm} \end{align}  

With the consensus condition~\eqref{eq:consensus_condition},  the safety constraints across all possible obstacle configurations hold for each trajectory:
\begin{equation}
 \min_{o^{(j_c)} \in \mathcal{O}} \mathbf{h}(\mathbf{s}^{(j)}_{t_0,[0,N_s]}, o^{(j_c)})\succ \mathbf{0},\quad  \forall j \in \mathcal{I}_0^{N_c-1},\ \forall j_c \in \mathcal{I}_0^{N_c-1}.\notag
\end{equation}
Therefore, the set  \( \mathcal{C}_{\text{cons}} \) is non-empty and forward invariant. As a result, the consensus trajectory segment from an initial safe set $\mathcal{C}_{\text{cons}} $ remains within the safety set $\mathcal{C}_{\text{cons}} $ over time evolution, i.e., $ \mathbf{s}^{(j)}_{t_0,[0,N_s]} \in  \mathcal{C}_{\text{cons}} \subseteq \mathcal{C}_{h}\Rightarrow \mathbf{s}^{(j)}_{t_k,[0,N_s]}\in  \mathcal{C}_{\text{cons}} \subseteq \mathcal{C}_{h}, \ \forall N_s < N$.
This completes the proof.
\end{IEEEproof}    

Note that the function \(\varphi(\cdot)\) extracts the position, velocity, acceleration, and orientation states of the EV to enforce consensus constraints, as discussed in Section~\ref{subsection:problem_statement}. Consequently, condition \eqref{eq:consensus_condition2} is satisfied, as safety interactions are governed by the position and heading angle of the EV, as demonstrated in \eqref{eq:polar_safety1} and \eqref{eq:polar_safety2}. This ensures that the proposed consensus barrier allows the EV to navigate safely, despite uncertainties in obstacle configurations.

\begin{remark} \label{remark2}
The safety associated with the shared consensus trajectory segment is valid across all considered obstacle configurations \(o^{(j)} \in \mathcal{O}\) for all \(j \in \mathcal{I}_0^{N_c-1}\). This ensures consistent risk coverage within this segment. Additionally, some trajectories may account for denser obstacle configurations, while others may have sparser configurations to allow for more proactive exploration. By sharing a common segment as a consensus constraint, all scenario-based trajectories maintain consistent risk coverage. This approach balances safety and efficiency, enabling the EV to complete its tasks without exhibiting overly conservative driving behaviors.  
\end{remark}
 
Note that we augmented the number of obstacles \(M_j\) to be the maximum number of obstacles \( M = \max\left\{ M_j \right\}_{j=1}^{N_c}\) observed across all obstacle configurations. This augmentation strategy can maintain consistent data structures across all configurations,  thus facilitating efficient parallel optimization in Section~\ref{subsec:CPTO}. For configurations with fewer obstacles, additional "virtual" obstacles are introduced. The safety ellipse parameters of virtual obstacles are set to zero (or near-zero values), ensuring they do not constrain the optimization process.  

\section{Parallel Consensus Optimization With Over-relaxed ADMM Iteration}
\label{subsec:CPTO}
In this subsection, we reformulate the initial motion planning problem \eqref{problem} into a bi-convex NLP problem, as described in Subsection \ref{subsec:problem_reformulation}. Subsequently, we decompose this bi-convex NLP problem into a series of lower-dimensional convex subproblems via consensus ADMM. This approach represents a specialized application of the Douglas-Rachford splitting method~\cite{eckstein1992douglas} to constrained optimization. The resulting framework efficiently solves each subproblem while rigorously enforcing consensus constraints through iterative updates.

% By leveraging consensus ADMM, we can efficiently solve each subproblem while enforcing the consensus constraints.
 \subsection{Problem Reformulation}   
 \label{subsec:problem_reformulation}
 Considering the spatiotemporal safety barrier \eqref{eq:polar_safety2} and the consensus constraints ~\eqref{eq:consensus_condition}, we can reformulate the planning problem \eqref{problem} into the following NLP problem to generate $N_c$ candidate trajectories in parallel:
    \begin{subequations}    
    \label{problem2}
      \begin{align}  
      \displaystyle\operatorname*{min}_{\substack{\{\mathbf{C}_x, \mathbf{C}_y,\mathbf{C}_\theta \}
     \\
    \{\bm{\omega^{(j)}}, \mathbf{d}^{(j)}\}}}~~ 
    \ &\ \sum_{j=0}^{N_c-1} \frac{1}{2}\mathbf{P}^T_{j}\mathbf{Q}_{P,j}\mathbf{P}_{j} 
   \label{eq:problem2_pto1}\\
    \operatorname*{s.t.}\quad\quad
    & \mathbf{A}_0 [  \mathbf{C}_x \quad  \mathbf{C}_y\quad \mathbf{C}_{\theta} ]  = \mathbf{E}_0,  \label{eq:problem2_pto2}\\ 
    &   \mathbf{A}_{f,xy}
    {[ \mathbf{C}_x\quad \mathbf{C}_y ] }=   \mathbf{E}_{f,xy},     \mathbf{A}_{f,\theta}    \mathbf{C}_{\theta}  = \mathbf{E}_{f,\theta} , \label{eq:problem2_pto3}\\ 
    &   \dot{\mathbf{W}}^T_B \mathbf{C}_{x}  - \mathbf{V} \cdot \cos{  (\mathbf{W}^T_B\mathbf{C}_{\theta})}  = \mathbf{0}, \label{eq:problem2_pto4}\\     
    &   \dot{\mathbf{W}}^T_B \mathbf{C}_{y}  - \mathbf{V} \cdot \sin{  (\mathbf{W}^T_B\mathbf{C}_{\theta})}  = \mathbf{0}, \label{eq:problem2_pto5}\\    
    & \mathbf{A}_h \mathbf{C}_{x}  = \mathbf{O}_{x}  + \mathbf{L}_x \cdot  \mathbf{D}    \cdot  \cos ( \bm{\omega}), \label{eq:problem2_pto6} \\
    & \mathbf{A}_h \mathbf{C}_{y} = \mathbf{O}_{y} +\mathbf{L}_y \cdot  \mathbf{D} \cdot 
 \sin ( \bm{\omega}),  \label{eq:problem2_pto7} \\  
   & \bm{\omega}^{(j)}   \in \mathcal{C}_{\omega}, \mathbf{d}^{(j)}  \in \mathcal{C}_d, \label{eq:problem2_pto8}\\
       & \mathbf{A}^T_{\text{cons},x} \mathbf{C}_x   = \mathbf{Y}_{x}, 
       \mathbf{A}^T_{\text{cons},y}  \mathbf{C}_y    = \mathbf{Y}_{y},\label{eq:problem2_pto9}  \\ 
      &  \mathbf{A}^T_{\text{cons},\theta}  \mathbf{C}_{\theta}     = \mathbf{Y}_{\theta} ,\label{eq:problem2_pto10}  \\
    &\mathbf{G} \mathbf{C}_x  -\mathbf{F}_x \preceq \mathbf{0},	\label{eq:problem2_pto11} \\
    &\mathbf{G} \mathbf{C}_y - \mathbf{F}_y  \preceq  \mathbf{0},\label{eq:problem2_pto12}\\ 
       &  \forall j \in \mathcal{I}_0^{N_c-1}. \notag
\end{align}
\end{subequations}  
Here, $\mathbf{Q}_{P,j} \in \mathbb{R}^{(n+1) \times (n+1)}$ is a weighting matrix to enhance the smoothness of all generated candidate trajectories in  B\'ezier curve form. The objective function  \eqref{eq:problem2_pto1} can be split into three separate terms as follows:
\begin{align} \vspace{-0mm}\label{eq:problem2_costfunc}
\begin{aligned}
      &\sum_{j=0}^{N_c-1} \frac{1}{2}\mathbf{P}^T_{j}\mathbf{Q}_{P,j}\mathbf{P}_{j} 
     = \frac{1}{2}\mathbf{P}^T\mathbf{Q}_{P}\mathbf{P}\\
    & =  \frac{1}{2} \mathbf{C}^T_x Q_{x}\mathbf{C}_x + \frac{1}{2} \mathbf{C}^T_y Q_{y}\mathbf{C}_y +  \frac{1}{2} \mathbf{C}^T_\theta Q_{\theta}\mathbf{C}_\theta\\
    & = f_x(\mathbf{C}_x) + f_y(\mathbf{C}_y) + g(\mathbf{C}_\theta),
\end{aligned}  \end{align} 
where \(\mathbf{C}_x\in \mathbb{R}^{(n+1)\times N_c}\), \(\mathbf{C}_y\in \mathbb{R}^{(n+1)\times N_c} \) and \(\mathbf{C}_\theta \in \mathbb{R}^{(n+1)\times N_c}\) denote the control point matrices to be optimized for generated trajectories in the longitudinal direction, lateral direction, and orientation aspects;  \([\mathbf{C}_x]_{i,j} = c^{(j)}_{x, i}\), \([\mathbf{C}_y]_{i,j} = c^{(j)}_{y, i}\), and \([\mathbf{C}_\theta]_{i,j} = c^{(j)}_{\theta, i}\).
\subsubsection{Boundary Constraints}
The constant matrices $\mathbf{A}_0 $, $\mathbf{A}_{f,xy} $ and  $\mathbf{A}_{f,\theta} $ are used to enforce the initial and terminal state conditions, defined as follows:
\begin{equation} \vspace{-0mm}
\mathbf{A}_0  =  \begin{bmatrix}  
    \mathbf{B}_{0,1}\quad \mathbf{B}_{1,1}\quad  \dots\quad \mathbf{B}_{n,1} \\
    \dot{\mathbf{B}}_{0,1} \quad\dot{\mathbf{B}}_{1,1} \quad \dots\quad \dot{\mathbf{B}}_{n,1} \end{bmatrix} \in \rr^{2\times (n+1)}, 
\notag
\vspace{-0mm}
\end{equation} 
 \begin{equation} \vspace{-0mm} 
 \mathbf{A}_{f,xy}  =  [\mathbf{B}_{0,N}\quad \mathbf{B}_{1,N}\quad  \dots\quad \mathbf{B}_{n,N}   ]  \in \rr^{n+1},
\notag
 \vspace{-0mm} \end{equation}  
 \begin{equation} \vspace{-0mm}\mathbf{A}_{f,\theta} = \begin{bmatrix}\mathbf{B}_{0,N}\quad \mathbf{B}_{1,N}\quad  \dots\quad \mathbf{B}_{n,N}   \\
\dot{\mathbf{B}}_{0,N} \quad\dot{\mathbf{B}}_{1,N} \quad \dots\quad \dot{\mathbf{B}}_{n,N} \end{bmatrix} \in \rr^{2\times (n+1)}.
\notag
  \vspace{-1mm} \end{equation} 
 
The matrix $\mathbf{E}_0\in \mathbb{R}^{6N_c}$ is introduced to enforce the initial position, velocity, heading angle, and yaw rate constraints for each candidate trajectory. Meanwhile, $\mathbf{E}_{f,xy}\in \mathbb{R}^{2N_c}$ guides trajectory optimization by sampling the target terminal position. Additionally, $\mathbf{E}_{f,\theta}\in \mathbf{0} \in \mathbb{R}^{2 N_c} $ ensures that the EV exhibits a small heading angle and yaw rate at the end of each planning horizon, facilitating stable driving.

\subsubsection{Safety Constraints}
The constraints \eqref{eq:problem2_pto6}-\eqref{eq:problem2_pto8} enforce the spatiotemporal safety constraint~\eqref{eq:polar_safety2}. $\mathbf{A}_h\in \mathbb{R}^{(N \times M \times N_c) \times (n+1)}$ denotes the augmented matrix $\mathbf{A}^{(j)}_h$ for all candidate trajectories, defined as follows:
\[
\mathbf{A}_{h} =   [\mathbf{A}^{(0)}_{h}  \quad \mathbf{A}^{(1)}_{h} \quad \cdots \quad \mathbf{A}^{(N_c-1)}_{h} ]^T, 
\] where \(\mathbf{A}^{(j)}_{h} = \mathbf{W}^T_B \otimes \mathbf{I}_{M} \in \mathbb{R}^{(N \times M) \times (n+1)}\) represents the vertical stacking of \(\mathbf{W}^T_B\) for \(M\) surrounding obstacles in obstacle configuration $o_j \in \mathcal{O}$. $\mathbf{L}_x \in \mathbf{l}^{(j)}_x \otimes \mathbf{I}_{M} \in \mathbb{R}^{(N \times M) \times N_c}$, $\mathbf{L}_y  = \mathbf{l}^{(j)}_y \otimes \mathbf{I}_{M}  \in \mathbb{R}^{(N \times M) \times N_c}$, where $\mathbf{l}^{(j)}_x \in \mathbb{R}^{N \times M}$ and $\mathbf{l}^{(j)}_y \in \mathbb{R}^{N \times M}$, with each term denoting the safe semi-major and semi-minor axes in safety ellipse along planning steps for the $j$-th trajectory.  These ellipses represent safety margins around the vehicle, ensuring that the planned trajectories maintain a safe distance from potential obstacles by adjusting their shape during planning. Similarly,  $\mathbf{D} = \mathbf{d}^{(j)} \otimes \mathbf{I}_{M} \in \mathbb{R}^{(N \times M) \times N_c}$,  $\bm{\omega}= \bm{\omega}^{(j)}\otimes \mathbf{I}_{M} \in \mathbb{R}^{(N \times M) \times N_c}$,  $\mathbf{O}_x = \mathbf{o}^{(j)}_x \otimes \mathbf{I}_{M}\in \mathbb{R}^{(N \times M) \times N_c}$,  $\mathbf{O}_y = \mathbf{o}^{(j)}_y \otimes \mathbf{I}_{M}\in \mathbb{R}^{(N \times M) \times N_c}$.
 
{Note that constraints \eqref{eq:problem2_pto6}-\eqref{eq:problem2_pto8} are bi-convex in the variables \( [\cos(\bm{\omega}),\, \sin(\bm{\omega})]^T \) and \( [\mathbf{C}_x,\, \mathbf{C}_y,\, \mathbf{D}]^T \), allowing decomposition into alternating convex subproblems during optimization.  }
The safety barrier set $\mathcal{C}_d$ for the scaling factor \({d}_k^{(j)} \) and the set $\mathcal{C}_{\omega}$ for $\omega^{(i)}_k$ are derived from the constraints (\ref{eq:polar_safety3}) and  (\ref{eq:angle_value}), respectively. 

\subsubsection{Consensus Constraints}
The constraints \eqref{eq:problem2_pto9}-\eqref{eq:problem2_pto10} are pivotal in ensuring that each candidate trajectory achieves consensus on specified trajectory variables, as elaborated in Sections~\ref{subsec:problem_reformulation} and \ref{subsubsec:consensus_barrier}. To uphold safety and driving consistency, consensus is enforced on the position, velocity, and acceleration in both longitudinal and lateral directions, as well as on the orientation state at the consensus segments. The associated consensus matrices \(\mathbf{A}_{\text{cons},x} \in \mathbb{R}^{((n+1) \times 3)  \times N_s}\), \(\mathbf{A}_{\text{cons},y} \in \mathbb{R}^{((n+1) \times 3)  \times N_s}\), and \(\mathbf{A}_{\text{cons},\theta}\in \mathbb{R}^{(n+1) \times N_s}\) are defined as follows: 
\[
\mathbf{A}_{\text{cons},x} =\mathbf{A}_{\text{cons},y} = \begin{bmatrix}  
    \mathbf{B}_{0,[1:N_s]}   & \mathbf{B}_{1,[1:N_s]} & \dots & \mathbf{B}_{n,[1:N_s]} \\
    \dot{\mathbf{B}}_{0,[1:N_s]} & \dot{\mathbf{B}}_{1,[1:N_s]} & \dots & \dot{\mathbf{B}}_{n,[1:N_s]} \\
    \ddot{\mathbf{B}}_{0,[1:N_s]} & \ddot{\mathbf{B}}_{1,[1:N_s]} & \dots & \ddot{\mathbf{B}}_{n,[1:N_s]}
\end{bmatrix} ,
\]  
\[
\mathbf{A}_{\text{cons},\theta} = \begin{bmatrix}  
    \mathbf{B}_{0,[1:N_s]} & \mathbf{B}_{1,[1:N_s]} & \dots & \mathbf{B}_{n,[1:N_s]} 
\end{bmatrix} .
\]
Here, \(\mathbf{B}_{i,[1:N_s]}\) denotes the sub-matrix comprising all rows and the first \(N_s\) columns of the matrix \(\mathbf{B}_i\) for each \(i \in \mathcal{I}^n_0\).
Each column of the consensus global variables \(\mathbf{Y}_{x} \in \mathbb{R}^{3N_s \times N_c}\), \(\mathbf{Y}_{y} \in \mathbb{R}^{3N_s \times N_c}\), and \(\mathbf{Y}_{\theta} \in \mathbb{R}^{N_s \times N_c}\) remains consistent across candidate trajectories to enforce the consensus constraint during iteration. 
\subsubsection{Motion Feasibility Constraints} 
The motion feasibility of the EV is governed by its nonholonomic kinematic constraints \eqref{eq:problem2_pto4}-\eqref{eq:problem2_pto5} throughout the planning horizon \(T\), where the matrix $\mathbf{V}$ is the vertical stacking of \(v^{(j)}_k\in \mathbb{R}^{N\times N_c}\) for all \(k \in \mathcal{I}_0^{N-1}\). 
Additionally, the EV should adhere to road boundaries and engine limitations. Hence, trajectory optimization needs to account for state constraints on position, acceleration, and jerk values, as constrained by \eqref{eq:problem2_pto11}-\eqref{eq:problem2_pto12}. The matrices \(\mathbf{G} \in \mathbb{R}^{6N\times (n+1)}\), and \(\mathbf{F}_x \in \mathbb{R}^{6N\times N_c}\) are defined as follows: 
\begin{align}
  \mathbf{G} = \begin{bmatrix}\mathbf{W}^T_B& -\mathbf{W}^T_B& \ddot{\mathbf{W}}^T_B& -\ddot{\mathbf{W}}^T_B& \dddot{\mathbf{W}}^T_B& -\dddot{\mathbf{W}}^T_B\end{bmatrix}^T,
  \notag
\end{align}
\begin{equation} \vspace{-0mm}
\mathbf{F}_x = [\mathbf{P}_{x,\max}\ -\mathbf{P}_{x,\min}\ {\mathbf{A}}_{x,\max}\ -{\mathbf{A}}_{x,\min}\ {\mathbf{J}}_{x,\max}\ -{\mathbf{J}}_{x,\min}]^T,
\notag
 \vspace{-0mm} \end{equation} 
 where 
\[
\begin{aligned}
    &\mathbf{P}_{x,\max}[k,j] = p_{x, \max}, \quad \mathbf{P}_{x,\min}[k,j] = p_{x, \min}, \\
    &\mathbf{A}_{x,\max}[k,j] = a_{x, \max}, \quad \mathbf{A}_{x,\min}[k,j] = a_{x, \min}, \\
    &\mathbf{J}_{x,\max}[k,j] = j_{x, \max}, \quad \mathbf{J}_{x,\min}[k,j] = j_{x, \min},
\end{aligned}
\] 
for all $ k \in \mathcal{I}_0^{N},  j \in \mathcal{I}_0^{N_c-1}$. Here, $ p_{x, \max}$ and $ p_{x, \min}$ denote the maximum and minimum longitudinal position limitations at time steps $k$, respectively;  $ a_{x, \max}$ and $ a_{x,\min}$ denote the maximum and minimum longitudinal acceleration limitations at time steps $k$, respectively; $ j_{x, \max}$ and $j_{x, \min}$ denote the maximum and minimum longitudinal jerk limitations at time steps $k$ for driving stability considerations, respectively. A similar definition applies to \(\mathbf{F}_y \in \mathbb{R}^{6N \times N_c}\) for the lateral states, capturing the constraints on lateral position, lateral acceleration, and lateral jerk.

To transform problem \eqref{problem2} into a standard ADMM form with feasibility guarantees, we need to address the inequality constraints \eqref{eq:problem2_pto11}-\eqref{eq:problem2_pto12} for further problem decomposition. Hence, we introduce two slack variables $\mathbf{Z}_x$ and $\mathbf{Z}_y$, resulting in:
\begin{subequations} \label{eq:transformed_ineq}
\begin{align}\vspace{-0mm}
    &\mathbf{G} \mathbf{C}_x  -\mathbf{F}_x + \mathbf{Z}_x = \mathbf{0}, \label{eq:transformed_ineq1}  \\
    &\mathbf{G} \mathbf{C}_y  -\mathbf{F}_y + \mathbf{Z}_y = \mathbf{0}.  \label{eq:transformed_ineq2} 
\end{align}\vspace{-0mm}
\end{subequations} 
Following this, we impose infinite penalties on the negative components of these slack variables by adding two terms $\mathcal{I}_{+}(\mathbf{Z}_x)$ and $\mathcal{I}_{+}(\mathbf{Z}_y)$ to the objective function \eqref{eq:problem2_costfunc}. 
% where the indicator function $\mathcal{I}_{+}(\cdot)$ is defined as follows:
% \[
% \mathcal{I}_{+}(\mathbf{Z}_x) = \begin{cases}
%     0, & \text{if } \mathbf{Z}_x \geq 0, \\
%     +\infty, & \text{otherwise}.
% \end{cases}
% \] 
 
\begin{remark}  
The objective function \eqref{eq:problem2_costfunc} consists of separable terms, combined with the separable nature of the constraints in \eqref{problem2}. This structure facilitates the decomposition of the bi-convex NLP problem \eqref{problem2} into a series of lower-dimensional convex subproblems, enabling real-time computation. By employing consensus ADMM, each subproblem can be solved in parallel while ensuring adherence to the consensus constraints~\eqref{eq:problem2_pto9}.
\end{remark}
 
\subsection{Problem Decomposition With Consensus ADMM}
\label{subsec:admm} 
\subsubsection{Augmented Lagrangian}
\label{subsubsec:lagrangian} 
We derive the \emph{augmented Lagrangian} of \eqref{problem2} for ADMM iteration, as follows:  
\begin{align} \vspace{-0mm} 
&\mathcal{L}\Big(  { \mathbf{C}_x, \mathbf{C}_y, \mathbf{C}_{\theta} }, {\mathbf{Z}_x, \mathbf{Z}_y},   {\bm{\omega}, \mathbf{D}},  \mathbf{Y}_x,\mathbf{Y}_y,\mathbf{Y}_{\theta} \nonumber\\
& \quad   \quad \bm{\lambda}_{x}, \bm{\lambda}_{y}, {\bm{\lambda}_{\theta}, \bm{\lambda}_{\text{obs},x}, \bm{\lambda}_{\text{obs},y}, \bm{\lambda}_{\text{cons},x}, \bm{\lambda}_{\text{cons},y}, \bm{\lambda}_{\text{cons},\theta}} 
\Big)  \nonumber\\
    &= f_x(\mathbf{C}_x) + f_y(\mathbf{C}_y) +g(\mathbf{C}_\theta) +  \mathcal{I}_{+}(\mathbf{Z}_x) + \mathcal{I}_{+}(\mathbf{Z}_y) \nonumber\\ 
    &\quad+ \bm{\lambda}^T_{x}\mathbf{C}_x  + \bm{\lambda}^T_{y}\mathbf{C}_y \nonumber \\
    &\quad+ \frac{\rho_{\theta}}{2}  \left\| \dot{\mathbf{W}}^T_B \mathbf{C}_{x}  - \mathbf{V} \cdot \cos{ (\mathbf{W}^T_B\mathbf{C}_{\theta})} 
    + \frac{\bm{\lambda}_{\theta} }{\rho_{\theta}}
    \right\|_{2}^{2} \nonumber\\
    &\quad+ \frac{\rho_{\theta}}{2}  \left\| \dot{\mathbf{W}}^T_B \mathbf{C}_{y}  - \mathbf{V} \cdot \sin{ (\mathbf{W}^T_B\mathbf{C}_{\theta})} 
    +\frac{\bm{\lambda}_{\theta} }{\rho_{\theta}}
    \right\|_{2}^{2} \nonumber \\
    &\quad+ \frac{\rho_{\text{obs}}}{2}  \left\| 
    \mathbf{A}_h \mathbf{C}_{x}  - \mathbf{O}_{x}  - \mathbf{L}_x \cdot  \mathbf{D}  \cdot  \cos ( \bm{\omega} ) 
    + \frac{ \bm{\lambda}_{\text{obs},x}}{\rho_{\text{obs}}} \right\|_{2}^{2} 
       \nonumber\\
    &\quad+ \frac{\rho_{\text{obs}}}{2}  \left\|    \mathbf{A}_h \mathbf{C}_{y}  - \mathbf{O}_{y} - \mathbf{L}_y \cdot  \mathbf{D}  \cdot  \sin ( \bm{\omega} )
    + \frac{ \bm{\lambda}_{\text{obs},y}}{\rho_{\text{obs}}} \right\|_{2}^{2} \nonumber\\ 
    &\quad+ \frac{\rho_{\text{cons},x }}{2}  \left\| \mathbf{A}^T_{\text{cons},x}  \mathbf{C}_x   - \mathbf{Y}_{x} 
     + \frac{ \bm{\lambda}_{\text{cons},x }}{\rho_{\text{cons},x}}  \right\|_{2}^{2} 
   \nonumber\\
     &\quad+ \frac{\rho_{\text{cons}, y }}{2}  \left\| \mathbf{A}^T_{\text{cons},y} \mathbf{C}_y    - \mathbf{Y}_{y} 
     + \frac{ \bm{\lambda}_{\text{cons},y }}{\rho_{\text{cons},y}}  \right\|_{2}^{2} 
   \nonumber\\
    &\quad+ \frac{\rho_{\text{cons},\theta }}{2}  \left\| \mathbf{A}^T_{\text{cons},\theta}  \mathbf{C}_{\theta}     - \mathbf{Y}_{\theta} 
      + \frac{ \bm{\lambda}_{\text{cons},\theta }}{\rho_{\text{cons},\theta}}  \right\|_{2}^{2} \nonumber\\ 
    &\quad+ \frac{\rho_x}{2} \left\| \mathbf{G} \mathbf{C}_x  -\mathbf{F}_x + \mathbf{Z}_x \right\|_{2}^{2} \nonumber\\
    &\quad+ \frac{\rho_y}{2} \left\| \mathbf{G} \mathbf{C}_y  -\mathbf{F}_y + \mathbf{Z}_y \right\|_{2}^{2}. 
    \label{lang_dual_problem}
\vspace{-0mm} \end{align} 
The structure of the \emph{augmented Lagrangian} function~\eqref{lang_dual_problem} allows us to group the primal variables into five categories: $\mathbf{C}_\theta$, $\{\mathbf{C}_x, \mathbf{Z}_x\}$, $\{\mathbf{C}_y, \mathbf{Z}_y\}$, $\bm{\omega}$, and $\mathbf{D}$ for optimization via consensus ADMM algorithm~\cite{boyd2011distributed}. The dual variables \( \bm{\lambda}_{\text{cons},x}  \in \rr^{3N_s\times N_c} \), \( \bm{\lambda}_{\text{cons},y}  \in \rr^{3N_s\times N_c} \)  and   \( \bm{\lambda}_{\text{cons},\theta}  \in \rr^{N_s\times N_c} \) are introduced to enforce the consensus constraints~\eqref{eq:problem2_pto9}-\eqref{eq:problem2_pto10}.  $\rho_{\text{cons},x}$, $ \rho_{\text{cons},y}$, $\rho_{\text{cons},\theta}$ are the corresponding $l_2$ penalty parameters and drive local variables toward their consensus global values $\mathbf{Y}_x$, $\mathbf{Y}_y$ and $ \mathbf{Y}_{\theta}$.  
$ \bm{\lambda}_{\theta}\in \rr^{N\times N_c}$, $ \bm{\lambda}_{\text{obs},x} \in \rr^{(N\times M)\times N_c} $, and $ \bm{\lambda}_{\text{obs},y} \in \rr^{(N\times M)\times N_c} $ are dual variables associated with the constraints \eqref{eq:problem2_pto4}-\eqref{eq:problem2_pto5} and \eqref{eq:problem2_pto6}-\eqref{eq:problem2_pto7}, respectively; 
The dual variables \( \bm{\lambda}_{x}  \in \rr^{(n+1)\times N_c} \)  and \( \bm{\lambda}_{y} \in \rr^{(n+1)\times N_c} \) are associated with inequality constraints \eqref{eq:problem2_pto11}-\eqref{eq:problem2_pto12} to enhance iteration stability, as outlined in~\cite{taylor2016training}; $\rho_{x}$, $\rho_{y}$, $\rho_{\theta}$, and $\rho_{\text{obs}}$ denote the corresponding $l_2$ penalty parameters.  

% \subsection{Optimization Procedure}  
\subsubsection{Primal Variables Update}
\label{subsubsec:primal_uodate} 
\begin{alignat}{2} \vspace{-0mm}
    \mathbf{C}^{\iota+1}_{\theta}  &:= \displaystyle\operatorname*{ \text{argmin}}_{\mathbf{C}_{\theta}}~~
    \mathcal{L}   \Big(  \{\mathbf{C}_\theta\}, \{ \mathbf{C}^{\iota}_x, \mathbf{Z}^{\iota}_x\}, \{ \mathbf{C}^{\iota}_y, \mathbf{Z}^{\iota}_y\},  \{\bm{\omega}^{\iota}\}, \{ \mathbf{D}^{\iota}\}, \nonumber \\
   &\qquad\qquad\qquad \{ \mathbf{Y}^{\iota}_x, \mathbf{Y}^{\iota}_y, \mathbf{Y}^{\iota}_{\theta}\}, \{ \bm{\lambda}^{\iota}_{\text{cons},x}, \bm{\lambda}^{\iota}_{\text{cons},y}, \bm{\lambda}^{\iota}_{\text{cons},\theta} \}, \nonumber \\ 
    &\qquad\qquad\qquad \{\bm{\lambda}^{\iota}_{\theta},  \bm{\lambda}^{\iota}_{x}, \bm{\lambda}^{\iota}_{y},  \bm{\lambda}^{\iota}_{\text{obs},x}, \bm{\lambda}^{\iota}_{\text{obs},y} \}   \Big) \label{eq:sub_qp1}\\
   &   \qquad \text{s.t.}\quad
    [\mathbf{A}_0 \quad \mathbf{A}_{f,\theta} ]^T \mathbf{C}_{\theta}  = [\bm{\theta}_0\quad \dot{\bm{\theta}}_0\quad \mathbf{0}]^T,  \nonumber 
 \end{alignat}
where $\bm{\theta}_0\in\mathbb{R}^{ N_c}$ and  $\dot{\bm{\theta}}_0\in\mathbb{R}^{ N_c}$ denote the initial heading angle and yaw rate for $N_c$ candidate trajectories, respectively;  $\iota$ denotes the current iteration number.
\begin{alignat}{2} \vspace{-0mm}
    \mathbf{C}^{\iota+1}_{x}  &:= \displaystyle\operatorname*{ \text{argmin}}_{\mathbf{C}_{x}}~~
    \mathcal{L}   \Big(  \{\mathbf{C}^{\iota}_\theta\}, \{ \mathbf{C}_x, \mathbf{Z}^{\iota}_x\}, \{ \mathbf{C}^{\iota}_y, \mathbf{Z}^{\iota}_y\},  \{\bm{\omega}^{\iota}\}, \{ \mathbf{D}^{\iota}\}, \nonumber \\
   &\qquad\qquad\qquad \{ \mathbf{Y}^{\iota}_x, \mathbf{Y}^{\iota}_y, \mathbf{Y}^{\iota}_{\theta}\}, \{ \bm{\lambda}^{\iota}_{\text{cons},x}, \bm{\lambda}^{\iota}_{\text{cons},y}, \bm{\lambda}^{\iota}_{\text{cons},\theta} \}, \nonumber \\ 
    &\qquad\qquad\qquad \{\bm{\lambda}^{\iota}_{\theta},  \bm{\lambda}^{\iota}_{x}, \bm{\lambda}^{\iota}_{y},  \bm{\lambda}^{\iota}_{\text{obs},x}, \bm{\lambda}^{\iota}_{\text{obs},y} \}   \Big) \label{eq:sub_qp2}\\
   &   \qquad \text{s.t.}\quad
   [ \mathbf{A}_0 \quad \mathbf{A}_{f,x} ]^T \mathbf{C}_{x}  = [\mathbf{P}_{x,0}\quad \dot{\mathbf{P}}_{x,0}\quad \mathbf{P}_{x,N} ]^T,    \nonumber 
\vspace{-0mm} \end{alignat}
where $\mathbf{P}_{x,0} \in \mathbb{R}^{ N_c}$ and  $\dot{\mathbf{P}}_{x,0}\in\mathbb{R}^{ N_c}$ denote the initial longitudinal position and velocity for $N_c$ candidate trajectories, respectively.   
\begin{align} \vspace{-0mm}
    \mathbf{C}^{\iota+1}_{y}  &:= \displaystyle\operatorname*{ \text{argmin}}_{\mathbf{C}_{y}}~~
    \mathcal{L}   \Big(  \{\mathbf{C}^{\iota}_\theta\}, \{ \mathbf{C}^{\iota}_x, \mathbf{Z}^{\iota}_x\}, \{ \mathbf{C}_y, \mathbf{Z}^{\iota}_y\},  \{\bm{\omega}^{\iota}\}, \{ \mathbf{D}^{\iota}\}, \nonumber \\
   &\qquad\qquad\qquad \{ \mathbf{Y}^{\iota}_x, \mathbf{Y}^{\iota}_y, \mathbf{Y}^{\iota}_{\theta}\}, \{ \bm{\lambda}^{\iota}_{\text{cons},x}, \bm{\lambda}^{\iota}_{\text{cons},y}, \bm{\lambda}^{\iota}_{\text{cons},\theta} \}, \nonumber \\ 
    &\qquad\qquad\qquad \{\bm{\lambda}^{\iota}_{\theta},  \bm{\lambda}^{\iota}_{x}, \bm{\lambda}^{\iota}_{y},  \bm{\lambda}^{\iota}_{\text{obs},x}, \bm{\lambda}^{\iota}_{\text{obs},y} \}   \Big) \label{eq:sub_qp3}
\end{align}
\begin{align}
       \qquad \text{s.t.}\quad
   [ \mathbf{A}_0 \quad \mathbf{A}_{f,y} ]^T \mathbf{C}_{y}  = [\mathbf{P}_{y,0}\quad \dot{\mathbf{P}}_{y,0}\quad \mathbf{P}_{y,N} ]^T,    \nonumber 
\vspace{-0mm} \end{align} where $\mathbf{P}_{y,0} \in \mathbb{R}^{ N_c}$ and  $\dot{\mathbf{P}}_{y,0}\in\mathbb{R}^{ N_c}$ denote the initial lateral position and velocity for $N_c$ candidate trajectories, respectively.  
 Note that $\mathbf{P}_{x,N} \in \mathbb{R}^{ N_c}$ and $\mathbf{P}_{y,N} \in \mathbb{R}^{ N_c}$ are target longitudinal and lateral position vectors for $N_c$ candidate trajectories, which are obtained using an adaptive sampling strategy detailed in Algorithm 1 of~\cite{zheng2024barrier}.  See Appendix for the analytical form of $ \mathbf{C}^{\iota+1}_{\theta}$, $ \mathbf{C}^{\iota+1}_{x}$, and $ \mathbf{C}^{\iota+1}_{y}$. 

 Referring to~\cite{ghadimi2015optimal}, the iteration for slack variables takes the following form to enforce the inequality constraints \eqref{eq:problem2_pto11}-\eqref{eq:problem2_pto12}:  
 \begin{subequations} \vspace{-2mm}
       \begin{align} 
       \mathbf{Z}^{\iota+1}_x = &\max  \Big( \mathbf{0},  \mathbf{F}_x - \mathbf{G}\mathbf{C}^{\iota+1}_x  \Big),   \label{eq:z_x_relaxedadmm_update} \\
       \mathbf{Z}^{\iota+1}_y = &\max  \Big( \mathbf{0},  \mathbf{F}_y - \mathbf{G}\mathbf{C}^{\iota+1}_y  \Big).   \label{eq:z_y_relaxedadmm_update}  
       \end{align}  
\end{subequations} 

Exploiting the condition \eqref{eq:angle_value} and \eqref{eq:polar_safety2}-\eqref{eq:polar_safety3} , we can get the analytical value for $\bm{\omega}$ and $\mathbf{D}$, as follows:  
\begin{equation} \vspace{-0mm}
    \bm{\omega}^{\iota+1} = \arctan\left(\frac{\mathbf{L}_x \cdot (\mathbf{A}_h \mathbf{C}^{\iota+1}_{y} - \mathbf{O}_{y})}{\mathbf{L}_y \cdot (\mathbf{A}_h \mathbf{C}^{\iota+1}_{x} - \mathbf{O}_{x})}\right),
    \label{eq:omega_admm_update}
 \vspace{-0mm} \end{equation}
{\begin{equation} \vspace{-0mm}
       \mathbf{D}^{\iota+1} = \max  \Big( \mathbf{1},   
  1 + (1-\bm{\alpha})\cdot(\mathbf{D}^{\iota} -1))\Big).
  \label{eq:d_admm_update}  \end{equation}}

 \subsubsection{Consensus Variables Update}
\begin{subequations} \vspace{-0mm}\small
    \begin{align}  
      &        \mathbf{Y}^{\iota+1}_{x} [:,j] =   \frac{1}{N_c} \sum_{i=0}^{N_c-1} \Big(\mathbf{A}^T_{\text{cons},x}  \mathbf{C}^{\iota+1}_x [:,i] + \frac{\bm{\lambda}^{\iota}_{\text{cons},x} [:,i]}{\rho_{\text{cons},x}}\Big) \label{eq:consensus_x_admm_update},   \\
      &              \mathbf{Y}^{\iota+1}_{y} [:,j] =   \frac{1}{N_c} \sum_{i=0}^{N_c-1} \Big(\mathbf{A}^T_{\text{cons},y}  \mathbf{C}^{\iota+1}_y [:,i] + \frac{\bm{\lambda}^{\iota}_{\text{cons},y} [:,i]}{\rho_{\text{cons},y}}\Big)  \label{eq:consensus_y_admm_update},\\
        &    \mathbf{Y}^{\iota+1}_{\theta} [:,j]  =    \frac{1}{N_c} \sum_{i=0}^{N_c-1}\Big(\mathbf{A}^T_{\text{cons},\theta}  \mathbf{C}^{\iota+1}_{\theta}  [:,i]  + \frac{\bm{\lambda}^{\iota}_{\text{cons},\theta} [:,i]}{\rho_{\text{cons},\theta}} \Big) \label{eq:consensus_theta_admm_update},
   \end{align}      
\end{subequations}  
for all $j \in \mathcal{I}^{N_c}_0$. 
\begin{remark}
  As elaborated in the consensus and sharing part of~\cite{boyd2011distributed}, the average dual variables for each consensus variable have an average value of zero after the first iteration, i.e., $\frac{1}{N_c} \sum_{i=0}^{N_c-1} \bm{\lambda}^{\iota}_{\text{cons},x} [:,i] = \mathbf{0}$. 
\end{remark} 

We can further simplify \eqref{eq:consensus_x_admm_update}-\eqref{eq:consensus_theta_admm_update} as:
 \begin{subequations} \vspace{-0mm}\small
    \begin{align}  
      &        \mathbf{Y}^{\iota+1}_{x} [:,j] =   \frac{1}{N_c} \sum_{i=0}^{N_c-1} \mathbf{A}^T_{\text{cons},x}  \mathbf{C}^{\iota+1}_x [:,i]  \label{eq:consensus_x_admm_updatesimp},  \quad \forall j \in \mathcal{I}^{N_c}_0, \\
      &              \mathbf{Y}^{\iota+1}_{y} [:,j] =   \frac{1}{N_c} \sum_{i=0}^{N_c-1}  \mathbf{A}^T_{\text{cons},y}  \mathbf{C}^{\iota+1}_y [:,i]    \label{eq:consensus_y_admm_updatesimp},\quad \forall j \in \mathcal{I}^{N_c}_0 ,\\
        &    \mathbf{Y}^{\iota+1}_{\theta} [:,j]  =    \frac{1}{N_c} \sum_{i=0}^{N_c-1} \mathbf{A}^T_{\text{cons},\theta}  \mathbf{C}^{\iota+1}_{\theta}  [:,i] , \quad \forall j \in \mathcal{I}^{N_c}_0  \label{eq:consensus_theta_admm_update_simp}.
   \end{align}      
\end{subequations}  
  
  \subsubsection{Dual Variables Update}  

 \begin{equation} \vspace{-0mm}
\bm{\lambda}^{\iota+1}_{\theta}  = \bm{\lambda}^{\iota}_{\theta} + \rho_{\theta}  \left(   \mathbf{W}^T_B\mathbf{C}^{\iota+1}_{\theta} -  
         \arctan \left(\frac{\dot{\mathbf{W}}^T_B  \mathbf{C}^{\iota+1}_{y} }{ \dot{\mathbf{W}}^T_B \mathbf{C}^{\iota+1}_{x} }\right) \right),\label{eq:lambda_theta_admm_update} 
 \vspace{-2mm} \end{equation}   
\begin{subequations} \vspace{-0mm}
       \begin{align} 
       \bm{\lambda}^{\iota+1}_x =&  \bm{\lambda}^{\iota}_x + \rho_{x} \mathbf{G}^T ( \mathbf{G} \mathbf{C}^{\iota+1}_x  -\mathbf{F}_x + \mathbf{Z}^{\iota+1}_x ),
\label{eq:lambda_x_admm_update} \\
        \bm{\lambda}^{\iota+1}_y = & \bm{\lambda}^{\iota}_y+ \rho_{y} \mathbf{G}^T ( \mathbf{G} \mathbf{C}^{\iota+1}_y  -\mathbf{F}_y + \mathbf{Z}^{\iota+1}_y ),
\label{eq:lambda_y_admm_update}
       \end{align}    \vspace{-\baselineskip}
\end{subequations}    
\begin{subequations}  \vspace{-2mm}
       \begin{align} 
       \bm{\lambda}^{\iota+1}_{\text{cons},x} =&  \bm{\lambda}^{\iota}_{\text{cons},x}  + \rho_{\text{cons},x}  (\mathbf{A}^T_{\text{cons},x}  \mathbf{C}^{\iota+1}_x  -\mathbf{Y}^{\iota+1}_{x} ),
\label{eq:lambda_consensusx_admm_update} \\
            \bm{\lambda}^{\iota+1}_{\text{cons},y} =&  \bm{\lambda}^{\iota}_{\text{cons},y}  + \rho_{\text{cons},y}  (\mathbf{A}^T_{\text{cons},y}  \mathbf{C}^{\iota+1}_y  -\mathbf{Y}^{\iota+1}_{y} ),
\label{eq:lambda_consensusy_admm_update} \\         \bm{\lambda}^{\iota+1}_{\text{cons},\theta} =&  \bm{\lambda}^{\iota}_{\text{cons},\theta}  + \rho_{\text{cons},\theta}  (\mathbf{A}^T_{\text{cons},\theta}  \mathbf{C}^{\iota+1}_{\theta}  -\mathbf{Y}^{\iota+1}_{\theta} ),
\label{eq:lambda_consensustheta_admm_update}
       \end{align}   \vspace{-\baselineskip}
\end{subequations}  
\begin{subequations}  \vspace{-2mm}
    \begin{align}   
        \bm{\lambda}^{\iota+1}_{\text{obs},x} &= \bm{\lambda}^{\iota}_{\text{obs},x} + \rho_{\text{obs}} \left( \mathbf{A}_h\mathbf{C}^{\iota+1}_{x} - \mathbf{O}_{x} \right. \notag \\
      &\quad\quad\quad\quad\quad \quad\quad \left. - \mathbf{L}_x \cdot  \mathbf{D}^{\iota+1} \cdot  \cos{(\bm{\omega}^{\iota+1})} \right) \label{eq:lambda_obsx_admm_update}, \\
        \bm{\lambda}^{\iota+1}_{\text{obs},y} &= \bm{\lambda}^{\iota}_{\text{obs},y} + \rho_{\text{obs}} \left(\mathbf{A}_h\mathbf{C}^{\iota+1}_{y} - \mathbf{O}_{y} \right. \notag \\
      & \quad\quad\quad\quad \quad\quad\quad \left. - \mathbf{L}_y \cdot  \mathbf{D}^{\iota+1} \cdot  \sin{(\bm{\omega}^{\iota+1})} \right) \label{eq:lambda_obsy_admm_update}.
   \end{align}    
\end{subequations}    
As outlined in~\cite{eckstein1994parallel,ghadimi2015optimal}, we employ over-relaxed iteration to update the dual variables to enhance iteration stability and streamline convergence speed for the inequality constraints \eqref{eq:problem2_pto11}-\eqref{eq:problem2_pto12}, as follows: \begin{subequations} \vspace{-0mm}
       \begin{align}  
 \bm{\lambda}^{\iota+1}_x = & \bm{\lambda}^{\iota}_x + \rho_{x} \mathbf{G}^T (  (1-\alpha_x)(\mathbf{Z}^{\iota+1}_x-\mathbf{Z}^{\iota}_x) \notag \\ 
  &+ \alpha_x( \mathbf{G} \mathbf{C}^{\iota+1}_x  -\mathbf{F}_x + \mathbf{Z}^{\iota+1}_x)  ), \label{eq:lambda_x_relaxedadmm_update}\\
  \bm{\lambda}^{\iota+1}_y = & \bm{\lambda}^{\iota}_y+ \rho_{y}  \mathbf{G}^T (
  (1-\alpha_y)(\mathbf{Z}^{\iota+1}_y-\mathbf{Z}^{\iota}_y)
   \notag \\ 
  &+ \alpha_y ( \mathbf{G} \mathbf{C}^{\iota+1}_y  -\mathbf{F}_y + \mathbf{Z}^{\iota+1}_y)). \label{eq:lambda_y_relaxedadmm_update}
       \end{align}  
\end{subequations}  
In the study, we select $1.5$ as the iteration coefficient for the relaxation parameters $\alpha_x$ and $\alpha_y$.

\subsubsection{Termination Criterion} 
 The iterative procedure is terminated when one of the following conditions is met. First, the primal residual meets a predefined threshold \(\epsilon^{\text{pri}}\), which is typically set between 0.1 and 1. Second, the iteration number $\iota$ reaches the maximum $\iota_{\max}$.  
\begin{remark}  
The generated candidate trajectories are continually updated by solving the optimization problem (\ref{eq:problem2_costfunc}) in a receding horizon manner. At each planning step, only the first step of the consensus trajectory segment is executed by the EV. By continuously integrating new sensory data, the EV replans each trajectory to reflect the latest state of the surrounding environments. This feedback receding horizon planning strategy allows the EV to adapt to rapidly changing environments.
\end{remark}

\section{{Experimental Results }}
\vspace{-0mm}
\label{sec:exp}   
 In this section, we validate our proposed CPTO for autonomous driving tasks under perception uncertainties using both synthetic dense obstacle environments and a real-world dataset from the Next Generation Simulation (NGSIM)\footnote{\url{https://data.transportation.gov/Automobiles/Next-Generation-Simulation-NGSIM-Vehicle-Trajector/8ect-6jqj}}. The experiments are conducted in partially observed environments to assess the effectiveness of our approach.

\vspace{-0mm} 
    \subsection{Simulation Setup}
    \label{sec:setup} 
    The experiments are run using C++ and ROS2 (Robot Operating System 2) on an Ubuntu 22.04 system, equipped with an AMD Ryzen 5 5600G CPU (six cores @ 3.90 GHz) and 16 GB of RAM. Visualization is performed using RVIZ in ROS2. The parameters for the experiments are configured based on \cite{ge2021numerically}, as shown in Table~\ref{table:Parameter_Settings}.  
      
   \begin{table}[tp]
    \centering
    \scriptsize
    \caption{{Parameters in the Driving Experiments}} \vspace{-0mm}
    \label{table:Parameter_Settings}
    \begin{tabular}{c c c}
    \hline
    Description         & Parameter with value   \\ \hline 
    Front axle distance to center of Mass &   $l_f = 1.06\,\text{m}$\\
    Rear axle distance to center of Mass &  $l_r =  1.85\,\text{m}$\\
    % Minimum safe ellipse parameters &  $l_{x,\min} = 6\,\text{m}$,  $l_{y,\min} = 4.5\,\text{m}$ \\
    Rear longitudinal perception range & $r_\text{rp} = 10\,\text{m}$\\
     Lateral perception range & $r_\text{lp} = 10\,\text{m}$\\
     Fully observed threshold & $s_d = 15\,\text{m}$\\
    Maximum anticipated number of obstacles & $M=5$\\
    Longitudinal position range  &  $p_{x} \in [-500\,\text{m}, 5000\,\text{m}] $\\
    Velocity range  &  $v \in [0\,\text{m/s}, 24\,\text{m/s}] $\\ 
    Longitudinal acceleration range  &  $a_{x} \in [-4\,\text{m/s}^2, 3\,\text{m/s}^2] $\\
    Lateral acceleration range  &  $a_{y} \in [-5\,\text{m/s}^2, 5\,\text{m/s}^2] $\\
    Longitudinal jerk range  &  $j_{x} \in [-6\,\text{m/s}^3, 6\,\text{m/s}^3] $\\
    Lateral jerk range  &  $j_{y} \in [-6\,\text{m/s}^3, 6\,\text{m/s}^3] $\\
    Maximum iteration number & $\iota_{\text{max}} = 200$\\
    $l_2$ penalty parameters &  $\rho_{\theta} = \rho_{x} = \rho_{y}  =5$, $ \rho_{\text{obs}} =6$ \\  
     &  $\rho_{\text{cons},x } =\rho_{\text{cons},y } = 4$, $\rho_{\text{cons},\theta} = 2$ \\
     Stopping criterion value for iteration & $\epsilon^{\text{pri}} = 0.1$\\
      Order of  B\'ezier curves  &  $n=10$ \\
    Smoothness weighing matrices &  $\mathbf{Q}_{P} = [100\quad 100\quad 150]^T$\\ 
    \hline 
    \end{tabular}\vspace{-0mm}
    \end{table} 
    
        \begin{table*}[t]
            \centering
            \scriptsize
            \caption{{Performance Comparison of Four Frameworks for High-Speed Navigation in Dense and Uncertain Obstacle Environments}}\vspace{-2mm}
            \label{tab:Static_results}    
           \begin{tabular}[c]{|c | *{2}{c} | *{1}{c} | *{4}{c} | *{1}{c} |}
                %% HEADER
                \hline
                \multirow{2}{*}{\textbf{Algorithm}} & 
                \multicolumn{2}{c|}{{\textbf{Safety}}} &
                \multicolumn{1}{c|}{\textbf{ {\textbf{Accuracy}}}} &
                \multicolumn{4}{c|}{{\textbf{Stability}}} &
                \multicolumn{1}{c|}{{\textbf{Opt. Time}}} \\
                  &\safety &  \mobility  & \efficiencycruise & \efficiencycomputation    \\
                \hline  
                \multirow{1}{*}{BPHTO}
                 & 3.16 &    6.01  & 0.0602 & {\textbf{3.0745}}  & {6.0153}  & 0.3013 &  1.2889  &   61.10    \\
                \hline           
                \multirow{1}{*}{Batch-MPC} 
                & 0.33  & 5.92 & 0.1864    & {35.5428}  & {15.9549} & \textbf{0.2399 }& \textbf{0.8748}
                 &  \textbf{38.65}  \\
                \hline
                \multirow{1}{*}{Control-Tree}
                &  2.50 &5.21 & 0.2329 & {63.6118} & {45.4532 }  & 0.6618 & 1.5754 
               & 220.39 \\  
                 \hline
                \multirow{1}{*}{{CPTO-0} }
                & {0.33}  & {5.80} & {\textbf{0.037}}  & {4.0216}& {5.0892}  & {0.4527} & {1.2991}
               &  {70.28}    \\         \hline
                    \multirow{1}{*}{\textbf{CPTO}}
                 &\textbf{0 }&\textbf{ 6.22 } &   0.0930 & {4.0066}  & {\textbf{5.0336}}  & 0.4202 & 1.1904 &72.88  \\
                %% END
                \hline
            \end{tabular}    \vspace{-0mm}
        \end{table*}

        \begin{table*}[t] 
            \centering
            \scriptsize
            \caption{{Performance Comparison of Three Frameworks in Cruising under Dense Traffic Using NGSIM Dataset}}   
            \vspace{-1mm}\label{tab:NGSIM_cruise_table_results}
           \begin{tabular}[c]{|c | *{1}{c} | *{1}{c} | *{5}{c} | *{1}{c} |}
                %% HEADER
                \hline
                \multirow{2}{*}{\textbf{Algorithm}} & 
                \multicolumn{1}{c|}{{\textbf{Safety}}} &
                \multicolumn{1}{c|}{\textbf{ {\textbf{Accuracy}}}} &
                \multicolumn{5}{c|}{{\textbf{Stability}}} &
                \multicolumn{1}{c|}{{\textbf{Opt. Time}}} \\
                &\safetyngsim &  \mobility  & \efficiencycruisengsim & \efficiencycomputation    \\
                \hline
                %% ENTRIES
                        %
                \multirow{1}{*}{BPHTO}
                  &6.73 & 0.1240 & {2.7889}  & {2.8450}  &0.2056 &0.5402 &0.0377 & 73.76  \\
                \hline           
                \multirow{1}{*}{Batch-MPC} 
              &  5.50  & 0.1601    &  {2.8838}  &  {17.3428 } &  \textbf{ 0.1874} 
               & \textbf{ 0.4109}    &0.0139& \textbf{48.87}  \\
                \hline
                \multirow{1}{*}{\textbf{CPTO}}
                &\textbf{7.02}  & \textbf{0.0493}  & {\textbf{2.1361}}  & {\textbf{2.7215}} &     0.1928 & 0.5903   &  \textbf{0.0098 }  &  75.41  \\
                % END
                \hline
            \end{tabular}    \vspace{-0mm}
        \end{table*}

  \subsubsection{{Scenarios and Parameters}} 
  \label{subsubsec:set_up}
At each time step, SVs (obstacles) are assumed to drive at a constant speed, introducing trajectory prediction errors for the EV. The position and velocity of the \(i\)-th SV are modeled as Gaussian distributions \cite{lefevre2014survey} for the EV to sample, as follows: 
\[
o_{x,k}^{(i)} \sim \mathcal{N}(0, \sigma_{x,k}^{(i)}),\quad o_{y,k}^{(i)} \sim \mathcal{N}(0, \sigma_{y,k}^{(i)}),
\] 
\[
v_{x,k}^{(i)} \sim \mathcal{N}(0, \sigma_{v_x,k}^{(i)}),\quad v_{y,k}^{(i)} \sim \mathcal{N}(0, \sigma_{v_y,k}^{(i)}),
\] 
where \(\sigma_{x,k}^{(i)}\), \(\sigma_{y,k}^{(i)}\), \(\sigma_{v_x,k}^{(i)}\), and \(\sigma_{v_y,k}^{(i)}\) represent the standard deviations of noise corresponding to the longitudinal and lateral position and velocity. These uncertainties are modeled as decreasing functions of the Euclidean distance \(s_{\text{dis}}\) between the EV and the obstacle. During simulations, the position noise is given by: 
\[
\sigma_{p_x,k}^{(i)} = \frac{\overline{\sigma}_{p_x}}{\max\left(\frac{10}{s_{\text{dis}} + 0.1}, 1\right)}, \quad     
\sigma_{p_y,k}^{(i)} = \frac{\overline{\sigma}_{p_y}}{\max\left(\frac{10}{s_{\text{dis}} + 0.1}, 1\right)},
\] 
where \(\overline{\sigma}_{p_x}\) and \(\overline{\sigma}_{p_y}\) are constant coefficients for longitudinal and lateral position noise, respectively. The same principle applies to velocity, with \(\overline{\sigma}_{v_x}\) and \(\overline{\sigma}_{v_y}\) representing the longitudinal and lateral velocity noise coefficients. In this study, we set \(\overline{\sigma}_{p_x}, \overline{\sigma}_{p_y}, \overline{\sigma}_{v_x}, \overline{\sigma}_{v_y}\) to $1\,\text{m}$, $0.5\,\text{m}$, $0.5\,\text{m/s}$, and $0.1\,\text{m/s}$, respectively. Additionally, the existence probability of obstacles becomes 1 once the distance between the EV and obstacles is less than a threshold $ s_{\text{dis}} < s_{e}$, where $s_{e}\sim \mathcal{N}(35\,\text{m}, 10\,\text{m})$. Note that once an obstacle is within a certain distance \(s_d\) of the EV, these standard deviations and existence uncertainties are reduced to zero, as specified in Assumption 1.  
 
The evaluation encompasses three typical and challenging driving scenarios, detailed in Sections \ref{subsec:comparative} and \ref{subsec:tradeoff}, which include: navigating at high speeds amidst dense and uncertain obstacle environments; {cruising in dense traffic conditions on freeway in the San Francisco Bay Area, utilizing the NGSIM dataset;} managing lane changing maneuvers in congested traffic scenarios, where SVs are controlled using the intelligent driver model (IDM). This model incorporates an additive white noise component with a variance of 0.2 in the acceleration control input. The planning steps in each horizon for the IDM and NGSIM datasets are set at $N=40$ and $N=50$, respectively, with an initial longitudinal velocity of $15\,\text{m/s}$ and a zero heading angle and acceleration across all scenarios.  The initial barrier coefficient parameter is set to 0.2, which linearly increases to 1 along the planning horizon. The safe semi-minor axis parameters are consistently set across all scenarios, with \(l_{y,\max} = 5.4\,\text{m}\) and \(l_{y,\min} = 4.5\,\text{m}\). The safe semi-major axis parameters are \(l_{x,\max} = 7.2\,\text{m}\) and \(l_{x,\min} = 6\,\text{m}\)  in Section~\ref{subsec:comparative}. All initial values of dual variables $ \bm{\lambda}_{\theta}, \bm{\lambda}_{\text{obs},x}, \bm{\lambda}_{\text{obs},y}, \bm{\lambda}_{\text{cons},x},\bm{\lambda}_{\text{cons},y},  \bm{\lambda}_{\text{cons},\theta}$  and $ \bm{\lambda}_{x}, \bm{\lambda}_{y}$ are set to zero.

     \subsubsection{Baselines and Evaluation Metrics}
   {We conduct an ablation study and evaluate the performance of the proposed CPTO against three state-of-the-art parallel trajectory optimization methods in dense obstacle environments for high-speed autonomous driving tasks.} The first baseline, Batch-MPC~\cite{adajania2022multi}, tailored for highway scenarios using Bergman alternating minimization, is configured using open-source code\footnote{\url{https://github.com/vivek-uka/Batch-Opt-Highway-Driving}}. The second method, BPHTO~\cite{zheng2024barrier}, is a parallel homotopic planning algorithm that employs over-relaxed ADMM iteration. The third method, Control-Tree~\cite{phiquepal2021control}, utilizes consensus ADMM optimization for autonomous driving under partial observability. We adopted its open-source implementation here\footnote{\url{https://github.com/ControlTrees/icra2021}}. {The last baseline  CPTO-0 is an ablated version of CPTO without consensus constraints \eqref{eq:problem2_pto9}-\eqref{eq:problem2_pto10}.} For all algorithms, we adjust parameters and set the same maximum iteration number to optimize parallel trajectories and ensure optimal performance.
   
   {Our evaluation metrics focus on four critical aspects: safety, quantified by the collision rate and mean absolute distance to the nearest obstacle; task accuracy, measured by the deviation from the desired longitudinal driving speed; driving stability, evaluated by acceleration, jerk, and yaw rate values; and computational efficiency, determined by average solving time. }

\subsection{Comparative Results}
\label{subsec:comparative}
\subsubsection{Scenario 1: Navigation Among Uncertain Dense Obstacles}
\label{subsubsec:static_env}
\begin{figure*}[tp]
	 	\centering  \hspace{-4mm}
                \subfigure[]{
			     \label{fig:IDM_Frenet-planner-demos}
			\includegraphics[scale=0.1455]{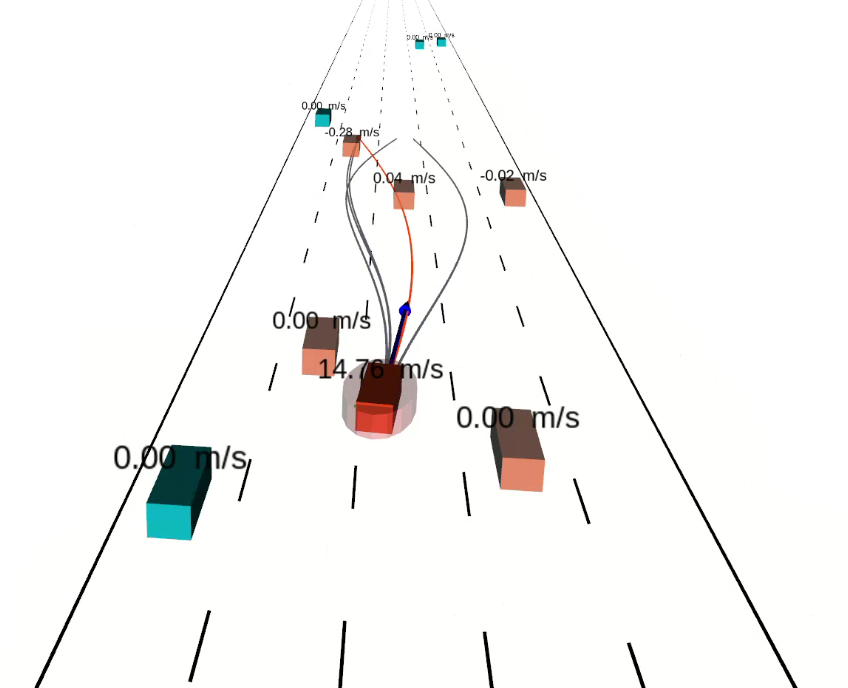}}\hspace{-4mm}
               \subfigure[]{
			\label{fig:IDM_ST-RHC-demos}
			\includegraphics[scale=0.1455]{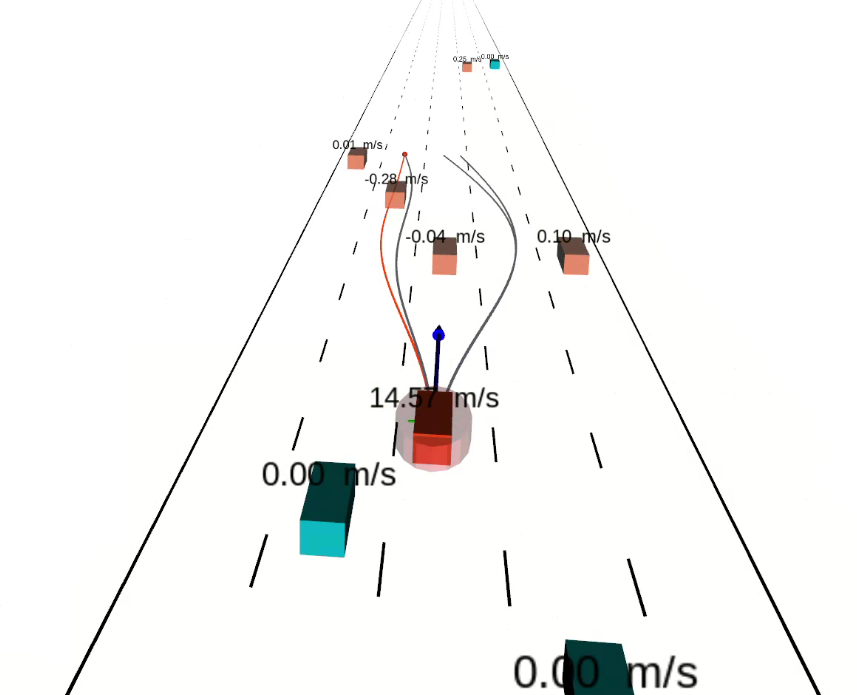}}\hspace{-4mm}
	 	\vspace{-0mm}  
            \subfigure[]{
			\label{fig:IDM_ST-RHC-demos}
			\includegraphics[scale=0.1455]{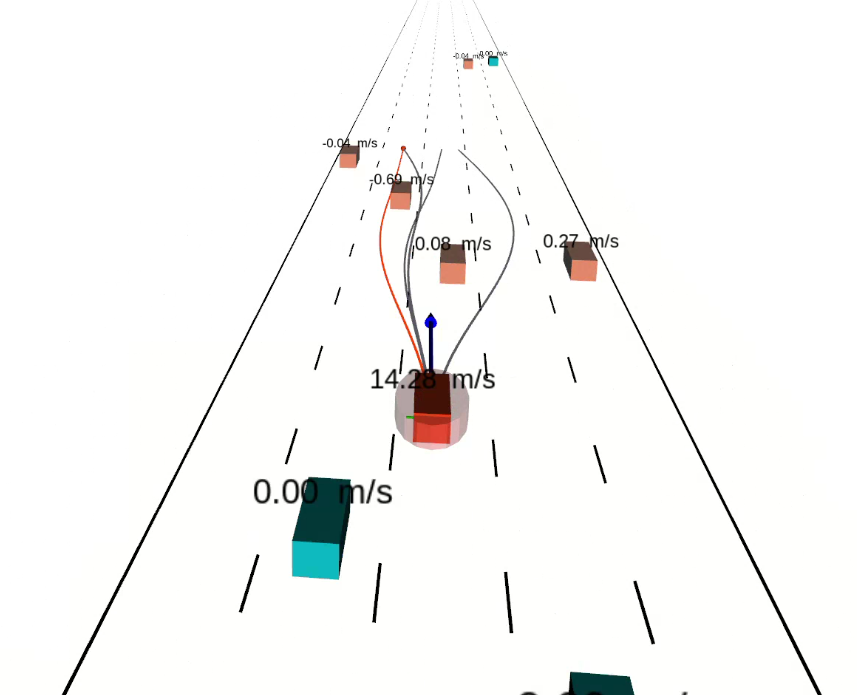}}\hspace{-4mm} 
            \subfigure[]{
			\label{fig:IDM_ST-RHC-demos}
			\includegraphics[scale=0.1455]{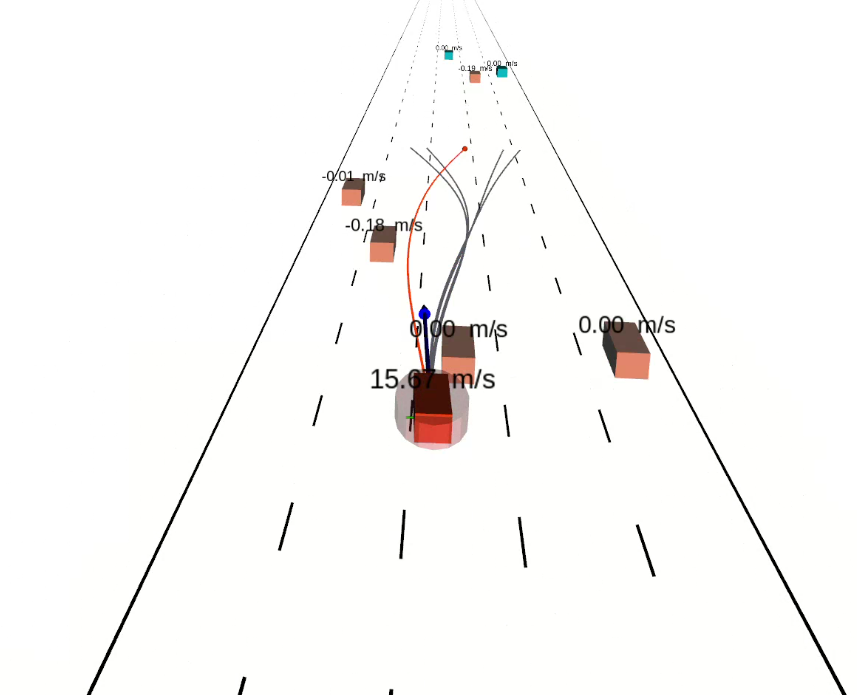}}\vspace{-2mm} \hspace{-4mm} \subfigure[ ]{
			     \label{fig:IDM_Frenet-planner-demos}
			\includegraphics[scale=0.1455]{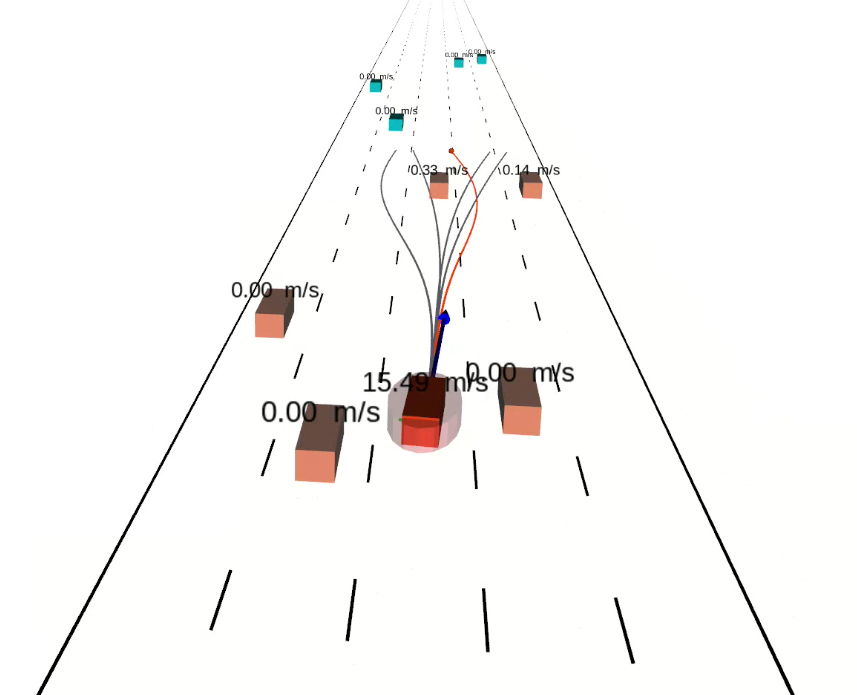}}\hspace{-4mm}
               \subfigure[]{
			\label{fig:IDM_ST-RHC-demos}
			\includegraphics[scale=0.1455]{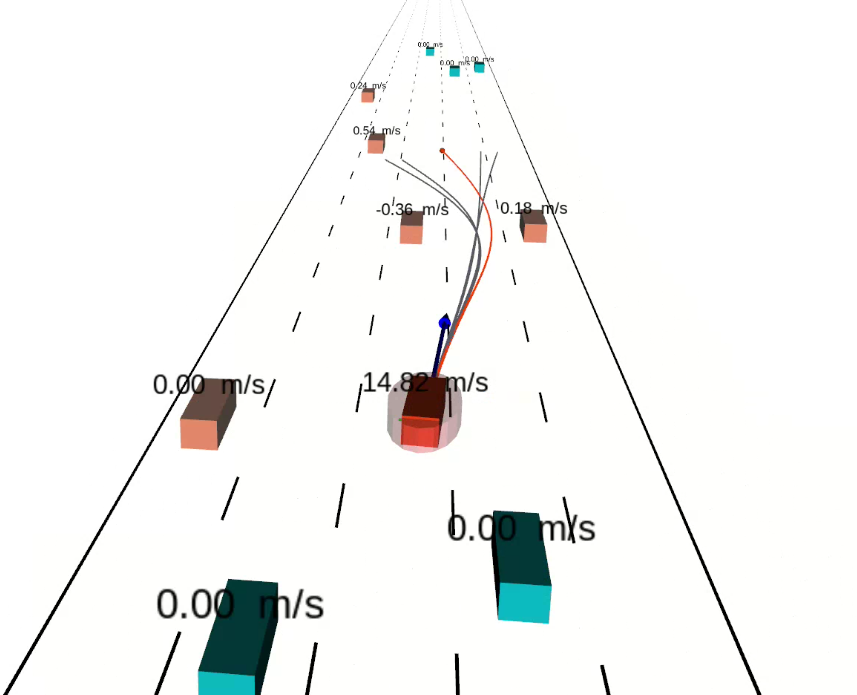}}\hspace{-4mm}
	 	\vspace{-0mm}  
            \subfigure[]{
			\label{fig:IDM_ST-RHC-demos}
			\includegraphics[scale=0.1455]{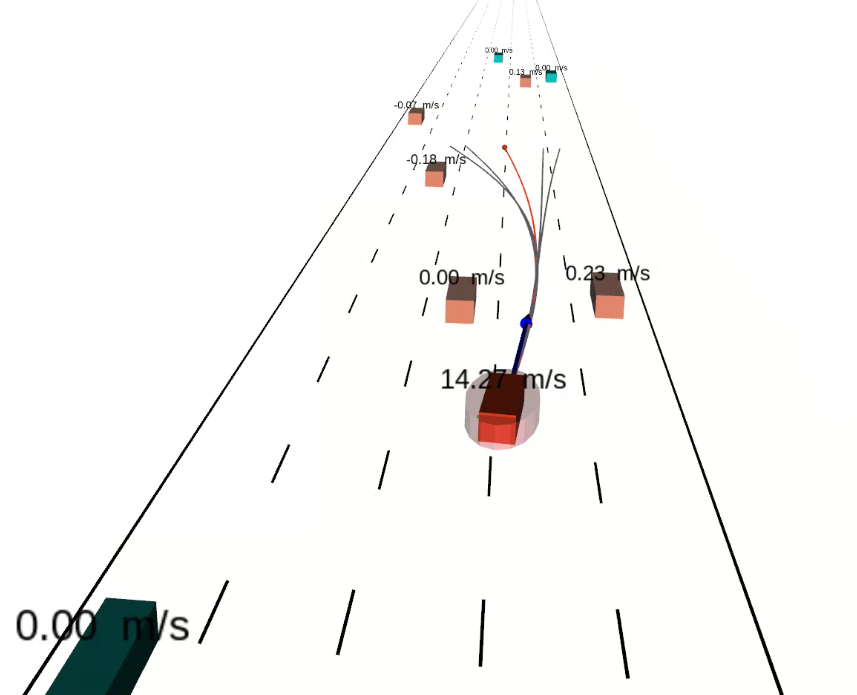}}\hspace{-4mm}  
            \subfigure[]{
			\label{fig:IDM_ST-RHC-demos}
			\includegraphics[scale=0.1455]{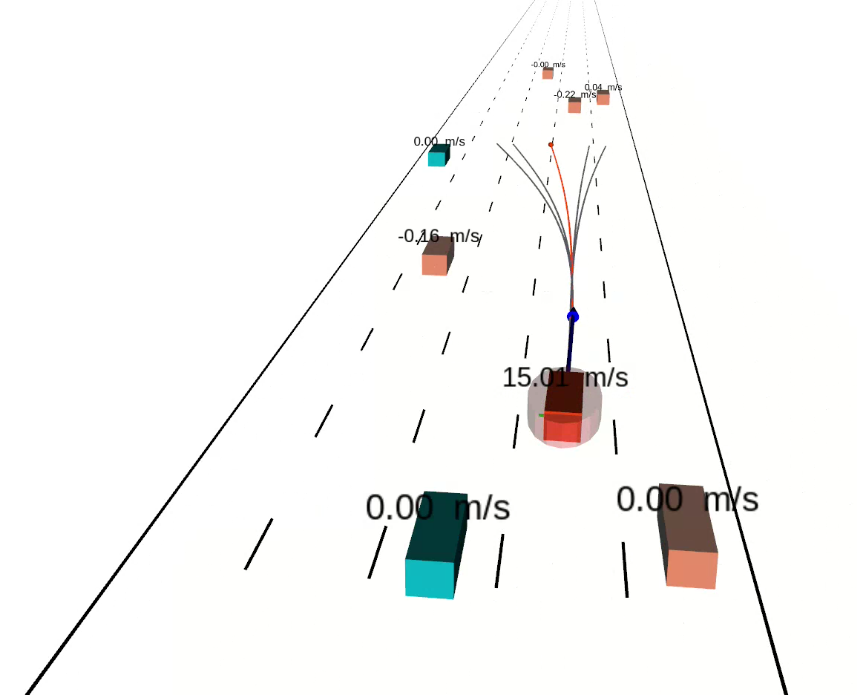}}\hspace{-1mm}
	 	\vspace{-0mm}
	 	\caption{Snapshots of the EV's trajectory in a dense obstacle environment under perception uncertainties. The EV, depicted by a red rectangle with a surrounding safe ellipsoid, optimizes several trajectories in parallel to navigate through obstacles.  Orange rectangles denote observed uncertain obstacles within the current planning framework, while blue rectangles denote other obstacles in the environment. The text above each rectangle shows the observed velocity of each obstacle. The blue arrow indicates the current velocity vector of the EV.} \vspace{-0mm}
	 	\label{fig:snapots_cruise_static}
	 	 	\vspace{-1mm}
	\end{figure*}

In this scenario, we address a common problem with false detections by sensors in autonomous driving, as outlined in~\cite{phiquepal2021control}. 
The EV is required to drive with an average longitudinal speed of $15\,\text{m/s}$ in the right four lanes, as depicted in Fig.~\ref{fig:snapots_cruise_static}. During the simulation, uncertain obstacles appear randomly.  On average, one potential obstacle appears every 15 meters, resulting in 60 uncertain obstacles encountered on a one-directional road with five driving lanes per minute. The initial position vector of the EV is set to $[-20\,\text{m}, -6\,\text{m}]^T$. 

The planning horizon is set to $4\,\text{s}$, with 10 steps per second. The number of candidate parallel trajectories to be optimized is $5$, with a consensus step $N_s=6$. Each trajectory corresponds to a different combination of object presences $o^{(j)}$, where $j \in \mathcal{I}^{5}_0$. In this simulation, the first, second, third, fourth, and last trajectories consider the number of nearest detected obstacles to be 2, 3, 3, 4, and 5, respectively. This strategy is designed to strike a balance between task accuracy and safety, as discussed in Remark 2. Note that the maximum anticipated number of obstacles considered during each planning period is 5, which is sufficient for most autonomous driving situations. 

    Table~\ref{tab:Static_results} presents the statistical results from 10 simulation runs, each with a duration of 600 steps. {It is evident that CPTO, CPTO-0, and BPHTO achieve relatively lower velocity tracking errors ${e}_{\text{mae}}$ compared to Control-Tree and Batch-MPC, indicating better driving accuracy.}
    { Although CPTO-0 achieves slightly better tracking accuracy than CPTO, it compromises safety performance and increases the risk of collisions, as indicated by its collision rate ($\mathcal{P}_{s} = 0.33\%$). Similarly, BPHTO, without trajectory coordination, exhibits a high collision rate ($\mathcal{P}_{s} = 3.16\%$) in this challenging high-speed scenario with uncertain dense static obstacles.} In contrast, CPTO achieves the lowest collision rate and the largest average distance ($\mathcal{S}_{\text{ma}} = 6.22\,\text{m}$) to the nearest obstacles among all algorithms. {These results demonstrate the advantages of the spatiotemporal consensus safety barrier module in CPTO, enabling high task accuracy and driving safety in high-speed driving scenarios with partially observed dense obstacles.}
 
    Regarding computational efficiency, one can notice that although Control-Tree uses multi-threading techniques for optimization, it necessitates a longer optimization time than other algorithms to handle dense obstacles. Consequently, its real-time performance cannot be achieved under dense obstacle environments ($\mathcal{T}_{\text{opt}} = 220.39\,\text{ms}$). {In contrast, the CPTO, CPTO-0, BPHTO, and Batch-MPC show significantly less optimization time than the Control-Tree approach,} supporting real-time replanning ($\mathcal{T}_{\text{opt}} < 100\,\text{ms}$) in dense obstacle environments. 
    
    % \textcolor{red}{The computation time of CPTO (72.88 ms) is moderately longer than both CPTO-0 (70.28 ms) and BPHTO (61.10 ms), primarily due to its additional consensus constraints for coordinated trajectory planning. While Batch-MPC achieves the fastest computation (38.65 ms) by focusing solely on control optimization, CPTO's marginally increased processing time represents a worthwhile trade-off for its integrated safety guarantees and stability enforcement.} 
  
    In terms of driving stability,  
    the maximum absolute longitudinal and lateral jerk values  ($\mathcal{J}_{x,\text{max}}$, $\mathcal{J}_{y,\text{max}}$) of CPTO and BPHTO are significantly lower than those of Batch-MPC and Control-Tree. This indicates that CPTO and BPHTO are capable of generating smoother longitudinal and lateral movements compared to Batch-MPC and Control-Tree in partially observable dense obstacle environments. {The reduced jerk enhances passenger comfort and maintains vehicle control stability during aggressive maneuvers.} While Batch-MPC exhibits smaller mean absolute acceleration values (\(\mathcal{A}_{x,\text{ma}}\), \(\mathcal{A}_{y,\text{ma}}\)) compared to the proposed CPTO, the latter achieves much smaller maximum absolute jerk values, demonstrating its ability to produce smoother acceleration and deceleration profiles. 

To better illustrate the navigation process under perception uncertainties, Fig.~\ref{fig:snapots_cruise_static} shows the trajectories and velocities of the EV during navigation. The top row (a-d) shows the EV's trajectory without consensus steps, where the EV switches between locally optimal candidate trajectories and collides with obstacles. {Similar phenomena can be observed for CPTO-0, BPHTO, and Batch-MPC algorithms with distinct optimized trajectories.} The bottom row (e-h) illustrates the EV's trajectory with 6 consensus steps, successfully avoiding collisions.  
    
\subsubsection{Scenario 2: Cruise under Dense Traffic with NGSIM Dataset}
\vspace{-0mm}
\label{subsubsec:cruise}
 This subsection evaluates the performance of the proposed CPTO in a cruising scenario under real-world dense traffic flow, where the motion of SVs is adopted from the real-world NGSIM dataset\footnote{\url{https://rb.gy/nr4yff}}. The data collection frequency of the NGSIM dataset is $12.5\,\text{Hz}$, ensuring a high temporal resolution. The simulation duration is configured as 450 steps, with a control and communication frequency of $12.5\,\text{Hz}$.  The initial position vector of the EV is $[-2\,\text{m}, -5\,\text{m}]^T$, with a target longitudinal velocity of $15\,\text{m/s}$.
 Other settings are the same as Section~\ref{subsubsec:static_env}. Due to the inability of the Control-Tree~\cite{phiquepal2021control} approach to effectively handle the dynamic dense traffic flow, its results are not included in this part. 
 
Table~\ref{tab:NGSIM_cruise_table_results} summarizes the performance of three algorithms under perception uncertainties. The average optimization time for all algorithms is less than $80\,\text{ms}$, ensuring real-time planning for the EV in dense traffic scenarios. Notably, the proposed CPTO algorithm achieves the lowest maximum cruise error $e_{\text{mae}}$ among the three algorithms, reducing the $e_{\text{mae}}$ by $69.2\%$ and $60.24\%$ compared to Batch-MPC and BPHTO, respectively. Furthermore, the average distance $\mathcal{S}_{\text{ma}}$ between the EV and SVs is larger with CPTO than with the other two algorithms. {These results indicate that CPTO can effectively handle the uncertain behaviors of SVs, while simultaneously maintaining safe operating distances and achieving precise tracking.}
% {Here, collision rates are omitted because SVs follow fixed historical trajectories and do not react to the EV (e.g., an SV will not slow down if the EV cuts into its path, potentially causing a collision). Such collisions reflect unavoidable historical interactions rather than algorithmic limitations; thus, $\mathcal{S}_{\text{ma}}$ is prioritized as a proactive safety metric for real-world-like non-reactive environments.}
% These results demonstrate that the CPTO with the consensus safety barrier module effectively handles the uncertain behaviors of SVs, leading to better task accuracy in real-time motion planning.
    \begin{figure}
		\centering
\includegraphics[width=8.5cm]{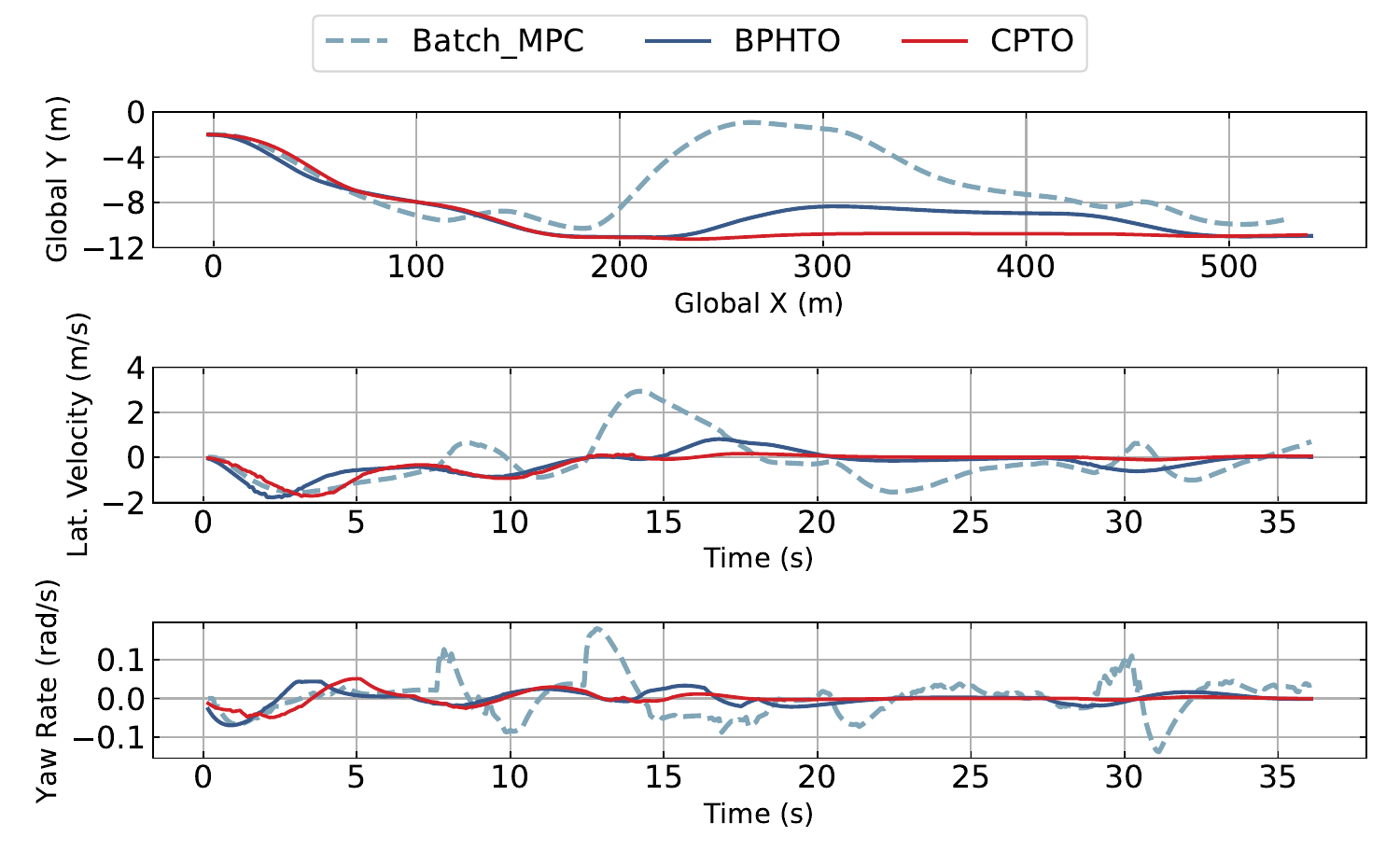}\vspace{-0mm}
		\caption{{Comparison of trajectory, lateral velocity, and yaw rate profiles of different algorithms when executing a cruising task using the NGSIM dataset.} } \vspace{-0mm}
  \label{fig:NGSIM_comparasion}
	\end{figure} 
 	\begin{figure}[tp]
	 	\centering  \hspace{1mm}
                \subfigure[ 16.2\,\text{s}]{
			     \label{fig:NGSIM_snapshot1}
			\includegraphics[scale=0.1905]{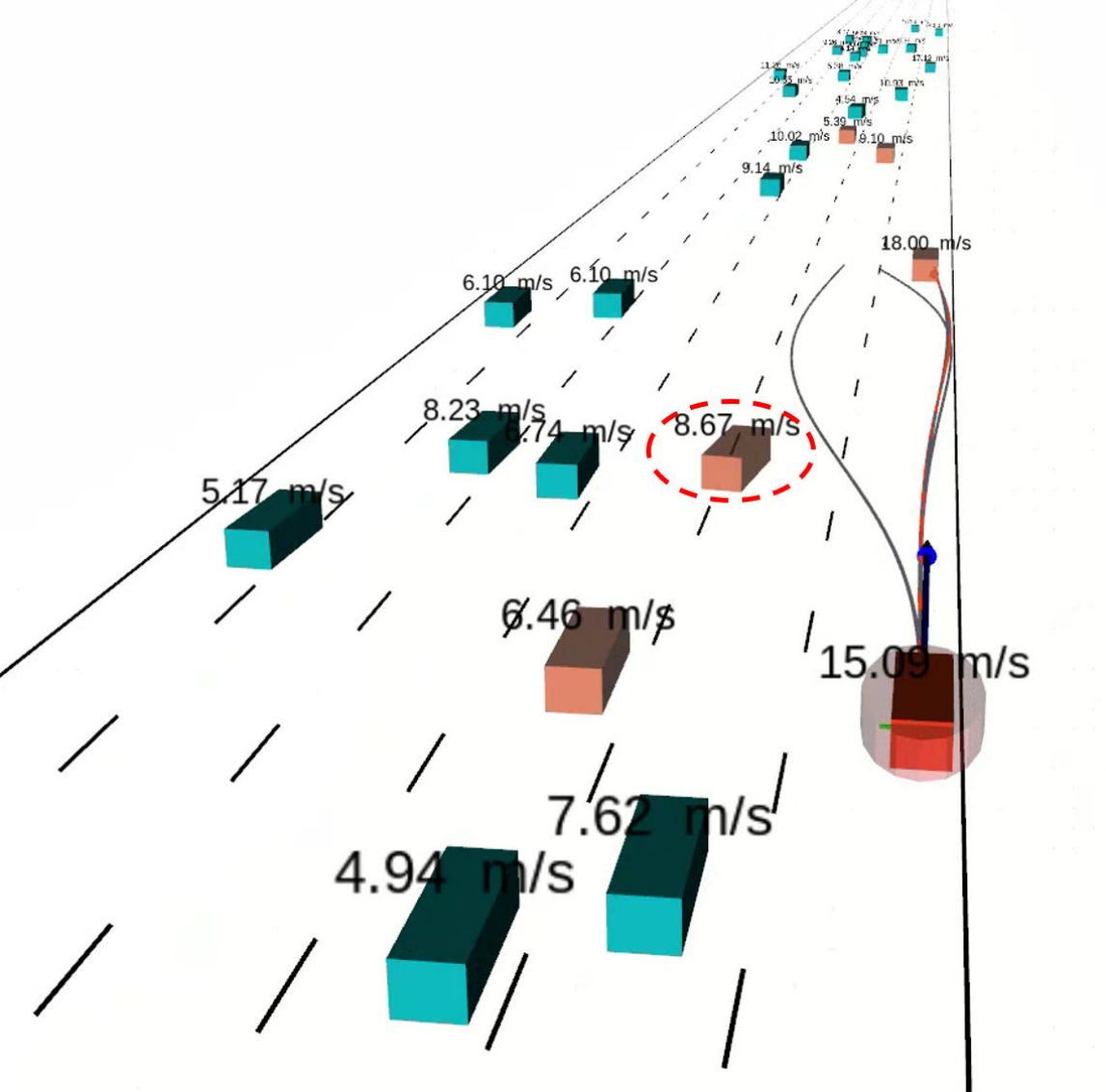}}\hspace{-0.5mm}
               \subfigure[17.5\,\text{s}]{
			\label{fig:NGSIM_snapshot2} \includegraphics[scale=0.1905]{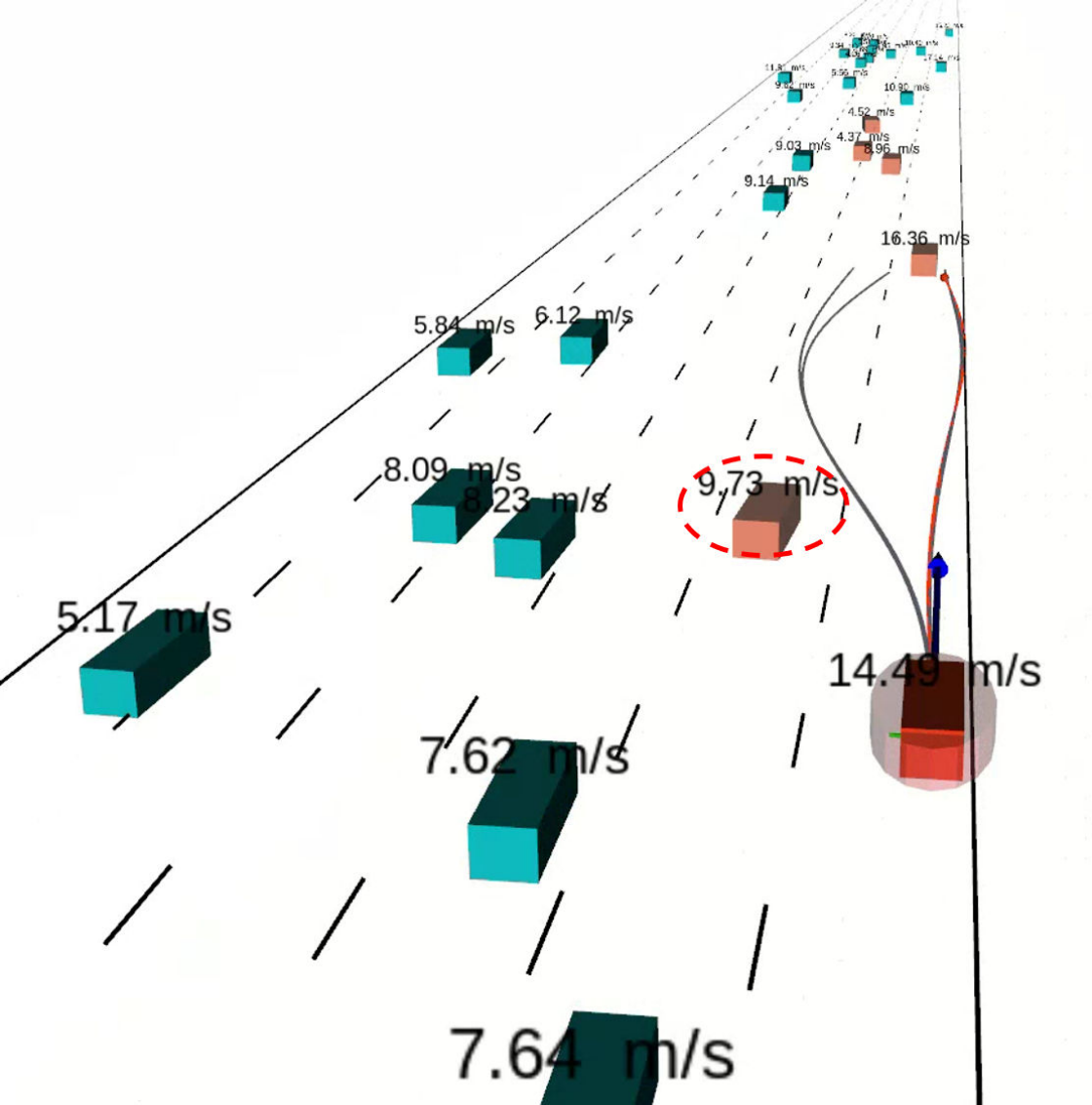}}
	 	\vspace{-1mm}  
          \hspace{1mm}  \subfigure[ 20.1\,\text{s}]{
			\label{fig:NGSIM_snapshot3}
			\includegraphics[scale=0.1905]{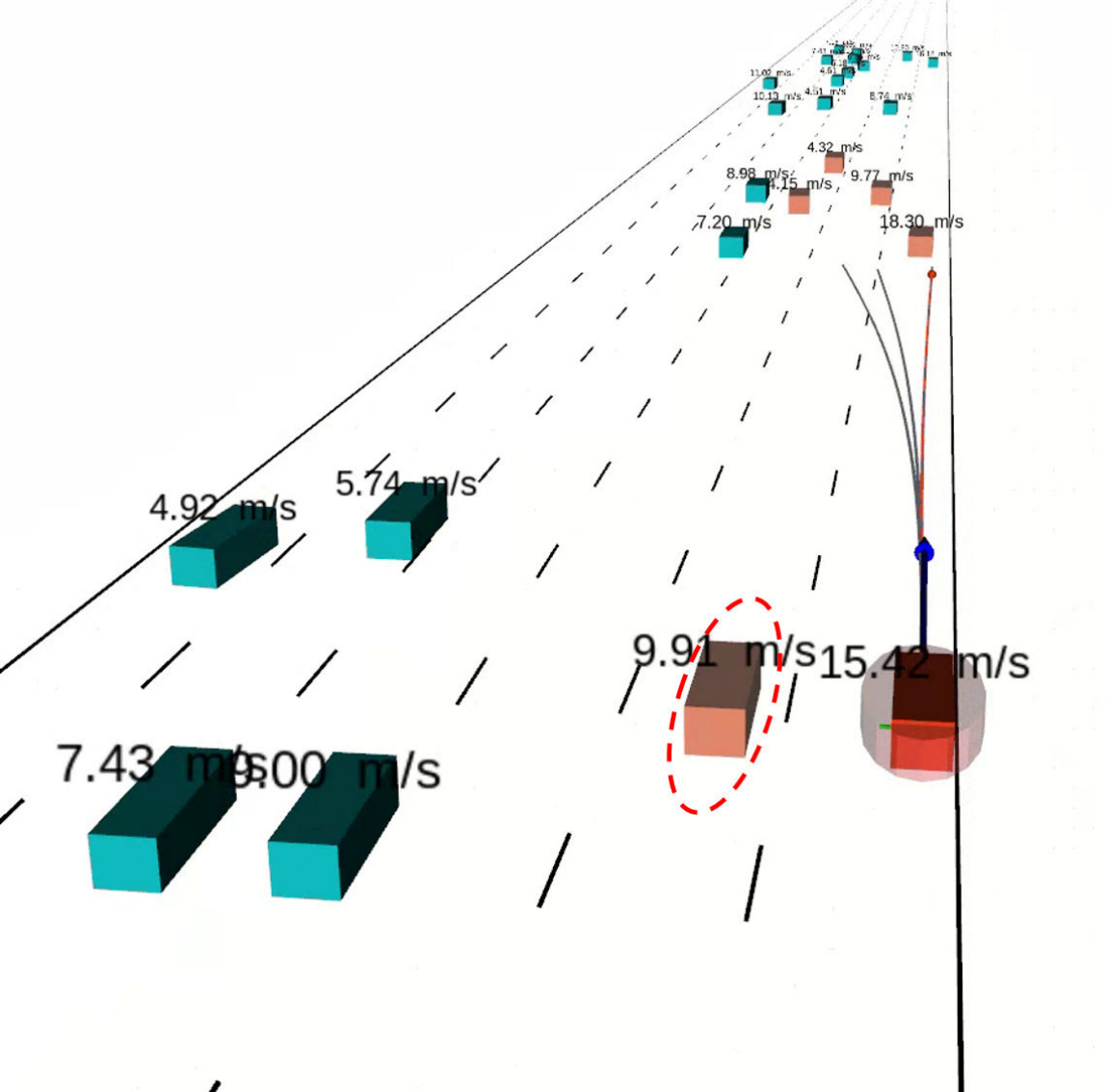}}\hspace{-0.5mm}  
            \subfigure[22.5\,\text{s}]{
			\label{fig:NGSIM_snapshot4}
			\includegraphics[scale=0.190]{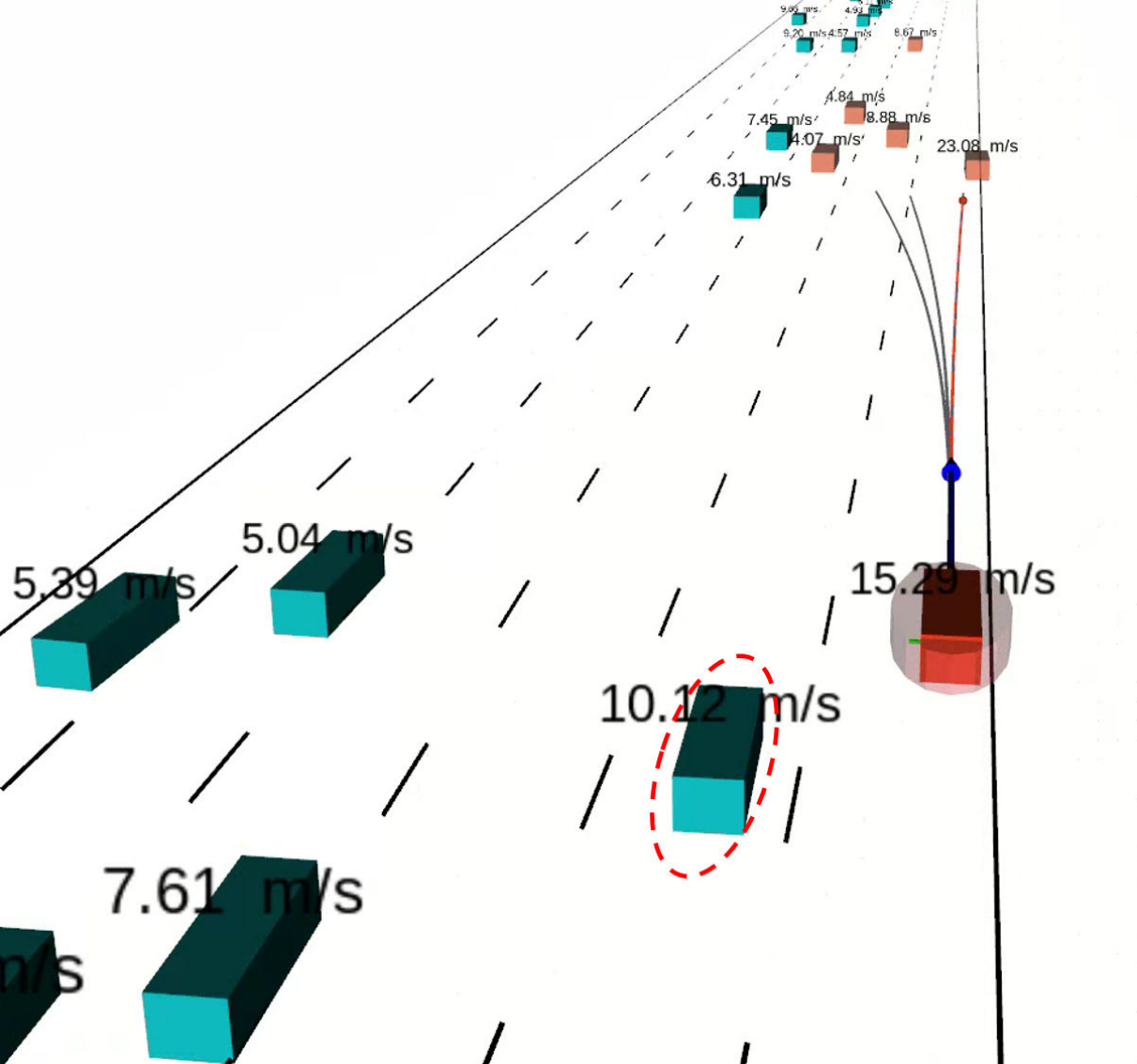}}\hspace{0mm}
	 	\vspace{-0mm}
	 	\caption{Snapshots of the EV's trajectory in a cruising scenario at 16.2\,\text{s}, 17.5\,\text{s}, 20.1\,\text{s} and  22.5\,\text{s} based on CPTO.  (a) and (b) illustrate that  the  EV yields to the non-cooperative lane-changing SV in the red dashed circle; (c) and (d)  illustrate that the EV accelerates to surpass this non-cooperative SV}		\vspace{-0mm}
	 	\label{fig:snapots_cruise_ngsim}
	 	 	\vspace{-0mm}
	\end{figure}	 
 
  In terms of driving stability, {Batch-MPC exhibits slightly better acceleration performance than CPTO, yet both significantly outperform BPHTO. Notably, CPTO achieves this comparable acceleration while maintaining substantially smoother jerk profiles.} One can notice that the maximum longitudinal jerk values $\mathcal{J}_{x,\text{max}}$ of CPTO and BPHTO are smaller than the value of Batch-MPC, indicating smoother longitudinal motion. {Specifically, CPTO reduces longitudinal jerk by $23.4\%$ compared to BPHTO and by $ 25.9\%$ versus Batch-MPC. Moreover, its maximum lateral jerk ($2.7215\,\text{m/s}^3$) is $84.3\%$ lower than Batch-MPC ($ 17.3428\,\text{m/s}^3$), significantly enhancing passenger comfort.}
  Furthermore, Fig.~\ref{fig:NGSIM_comparasion} shows that Batch-MPC achieves the worst driving stability among the three algorithms. This is evident from its trajectory, lateral velocity, and yaw rate evolution profiles, which exhibit significant fluctuations. These instabilities correspond to a 6.4 times higher maximum lateral jerk ($17.3428 \text{m/s}^3$) in Batch-MPC versus CPTO, potentially causing lateral control challenges during aggressive maneuvers. Notably, the yaw rate of Batch-MPC varies considerably, leading to larger changes in absolute lateral velocity compared to CPTO and BPHTO. These observations are further supported by the maximum lateral jerk value $\mathcal{J}_{y,\text{max}}$ ($17.3428\,\text{m/s}^3$) of Batch-MPC, which is much higher than those of CPTO and BPHTO. In contrast, CPTO exhibits the smallest lateral velocity changes among all algorithms. {Its average yaw rate ($0.0098\,\text{rad/s}$) is significantly smaller than those of BPHTO and Batch-MPC, enabling more stable lateral maneuvers and improved driving consistency that enhance passenger comfort.} These findings demonstrate that CPTO generates more stable driving behaviors than the other two algorithms, with enhanced safety further quantified by the mean absolute safety distance.  

To gain a more intuitive understanding of the capability of the proposed CPTO in handling dynamic obstacles under perception uncertainties, Fig.~\ref{fig:snapots_cruise_ngsim} illustrates a typical interaction process where the proposed CPTO handles a suddenly appearing, non-cooperative SV. At the time instant $16.2\,\text{s}$, the red EV, cruising at a speed of $15.09\,\text{m/s}$, detects a lane-changing SV. In response, the EV executes a slight deceleration maneuver to $14.49\,\text{m/s}$ to avoid a potential collision, as shown in Fig.~\ref{fig:NGSIM_snapshot2}. When the EV gets closer to the SV with reduced perception noise and detects that the SV does not perform consecutive lane-changing behaviors, the EV accelerates to quickly surpass this SV and then reduces its speed towards its target cruise speed, as depicted in Fig.~\ref{fig:NGSIM_snapshot3} and Fig.~\ref{fig:NGSIM_snapshot4}. This process can also be observed through the motion trajectory, lateral velocity, and yaw rate from $15\,\text{s}$ to $23\,\text{s}$, as illustrated in Fig.~\ref{fig:NGSIM_comparasion}. Throughout this entire process, all generated trajectories of the CPTO share a common trajectory segment to facilitate driving consistency under perception uncertainties.
      
  \begin{figure}
		\centering
		\includegraphics[width=8cm]{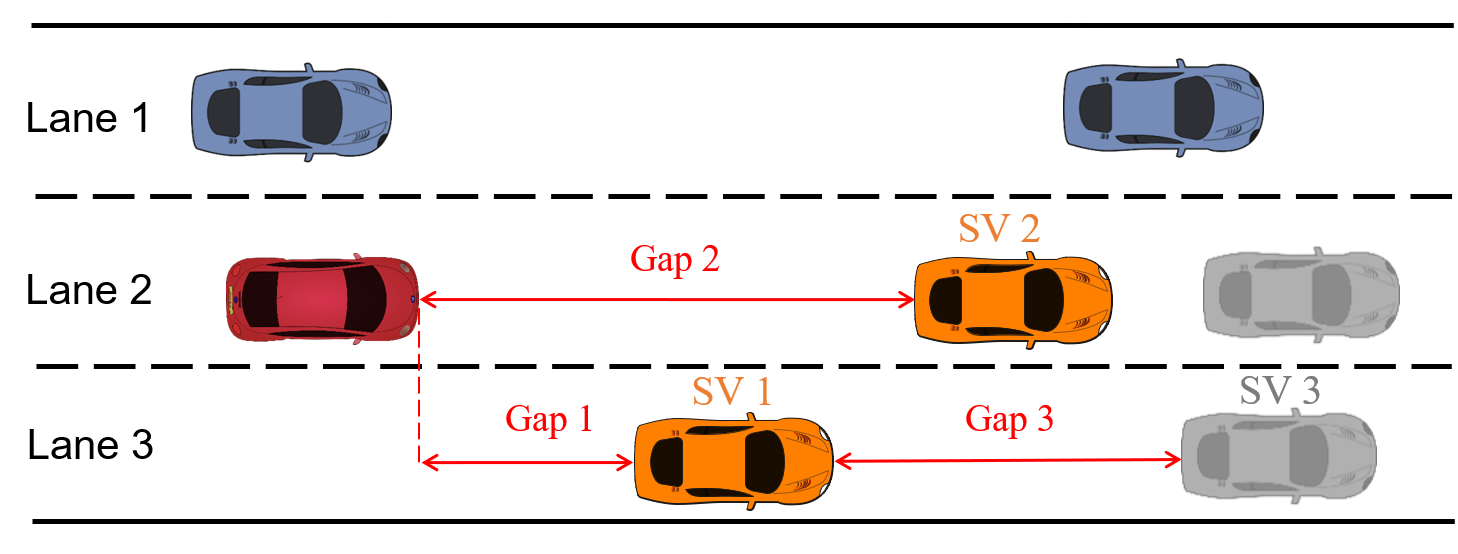} \vspace{-0mm}
		\caption{Illustration of the lane changing problem under dense traffic flow and perception uncertainties. The red EV in lane 2 is executing a lane change to lane 3. During this process, three gaps are available for the EV to choose from. The EV can choose to change to the rear or front of the SV1.} \vspace{-0mm}
  \label{fig:Lane_changing_diagram}
	\end{figure} 
 \begin{table}[t]
            \centering
            \scriptsize
            \caption{{Performance Comparison Among Four Algorithms in a Lane Changing Task under Dense Traffic}}   \vspace{-1mm}
            \label{tab:merge_results}    
            \begin{tabular}[c]{|c |  *{1}{c} |  *{2}{c} | *{1}{c}  |}
                %% HEADER
                \hline
                \multirow{2}{*}{\textbf{Algorithm}} &  
                \multicolumn{1}{c|}{\textbf{ Safety}} &
                \multicolumn{2}{c|}{\textbf{ Stability}} &
                \multicolumn{1}{c|}{\textbf{ Solving Time}}  \\
                  & \safetymerge & \mergeconsistency & \efficiencycomputation  \\
                \hline
                %% ENTRIES  
                \multirow{1}{*}{BPHTO }
               & 9.8486 &0.8797   & 0.0225 &\textbf{ 32.28 }  \\
                \hline     
                               \multirow{1}{*}{{CPTO-0}}
               & 13.3697  & 0.4999  & 0.0165 &  34.25   \\
                \hline     
                \multirow{1}{*}{CPTO-8}
                & \textbf{13.6259 } & 0.4772 & \textbf{0.0147} & 34.80 \\     
                 \hline
                \multirow{1}{*}{CPTO-15}
                &  13.2591  & \textbf{0.4626} & 0.0148 & 35.64\\
                %% END
                \hline
            \end{tabular}    \vspace{-1mm}
        \end{table}
 
      \begin{figure}
		\centering
		\includegraphics[width=8.5cm]{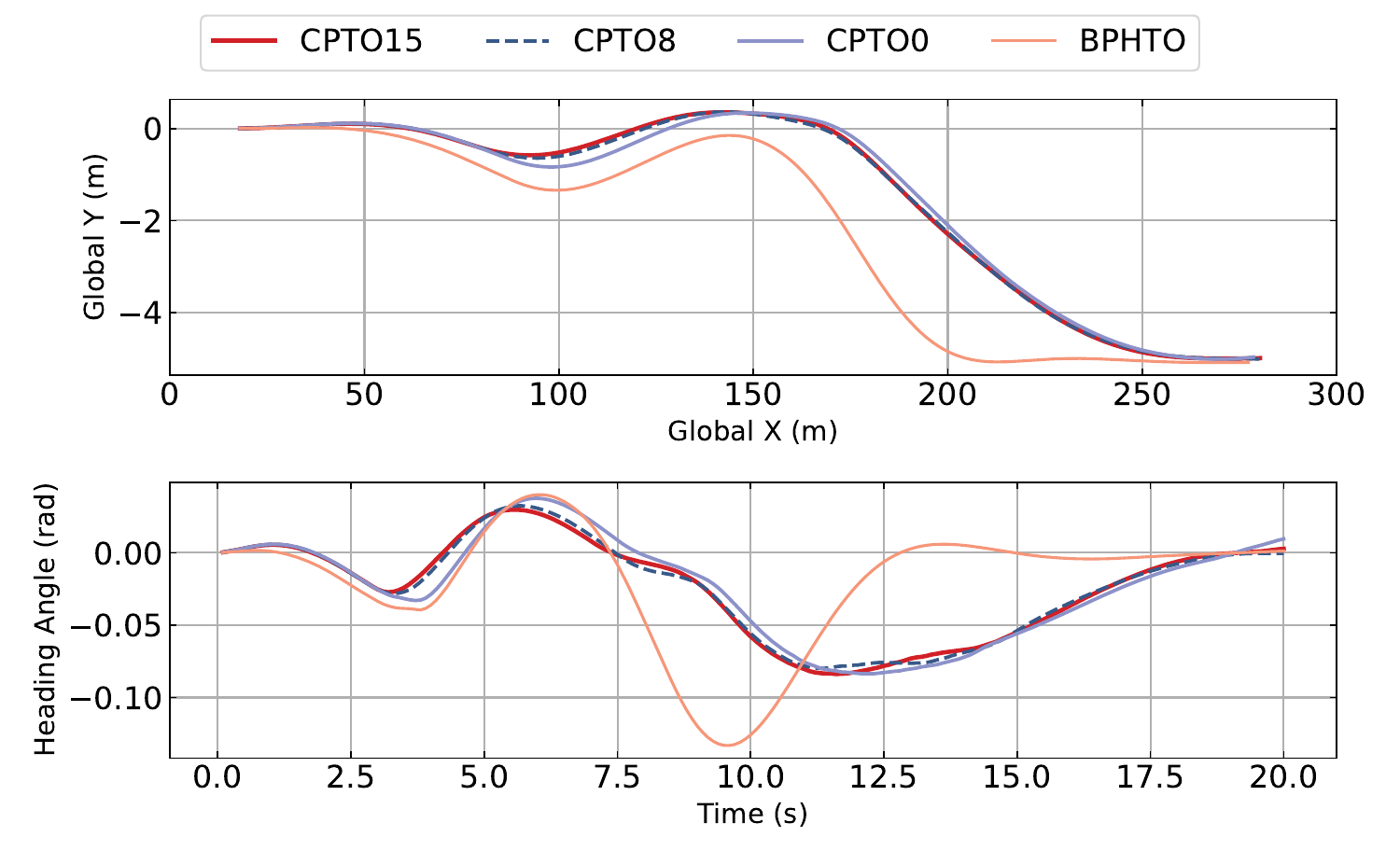}\vspace{-2mm}
		\caption{ Comparison of trajectory and heading angle profiles when executing a lane changing task under dense traffic, where SVs are controlled using IDM.} \vspace{-0mm}
  \label{fig:Merge_comparasion}
	\end{figure} 

     \begin{figure}[t]
		\centering		\includegraphics[width=8.5cm]{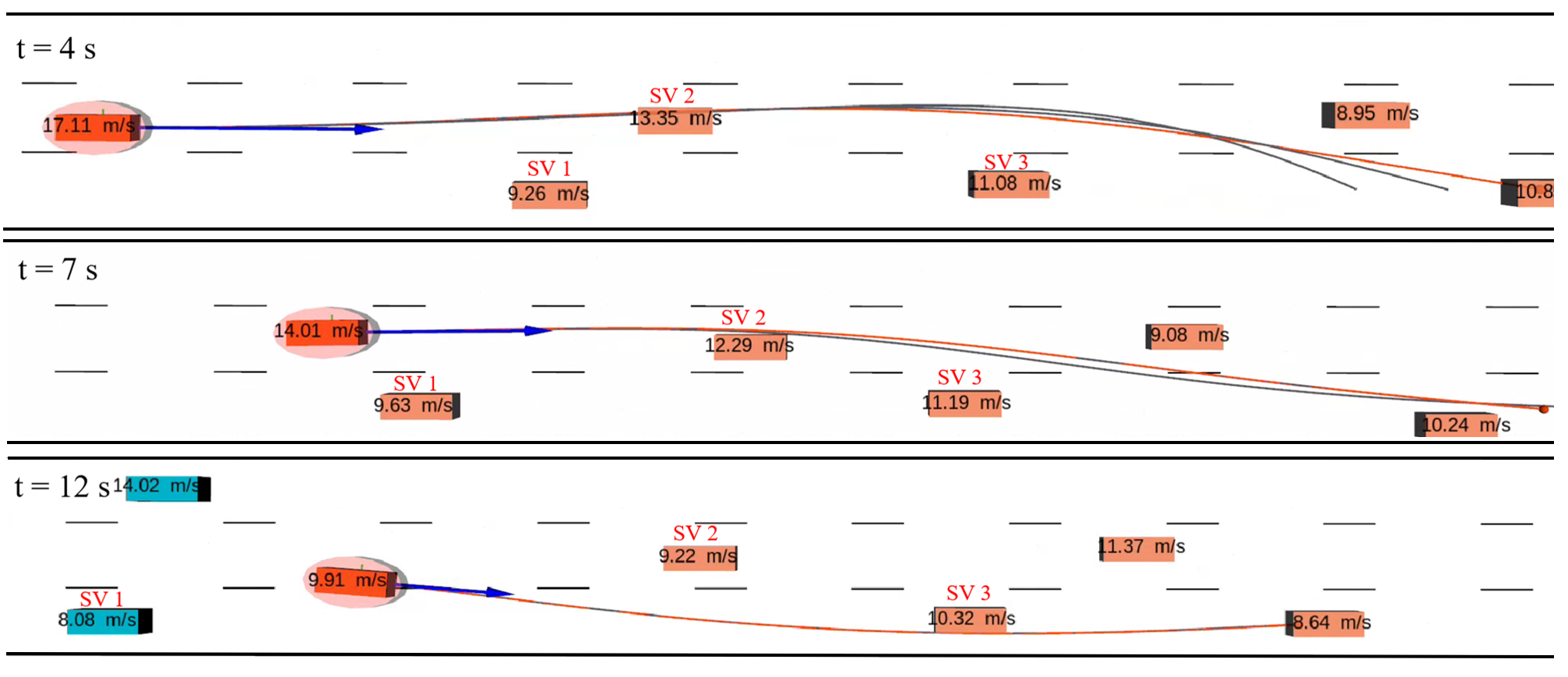}\vspace{-1mm}
		\caption{ Snapshots of the EV’s trajectory over the planning horizon ($T = 4\,\text{s}$) in a lane changing scenario at 4\,\text{s}, 7\,\text{s}, 12\,\text{s} based on CPTO planner with 15 consensus steps. The red EV dynamically adjusts its speed and orientation to execute a lane-changing behavior to the bottom lane under perception uncertainties. } \vspace{-0mm}
  \label{fig:snapshot_merge}
	\end{figure}  
    \section{{Discussion}}

\begin{table}[t] 
\centering
\scriptsize
\caption{{Computation Time (ms) Across Varying Number of Trajectories and Obstacles}} 
\begin{minipage}{0.45\textwidth}
\centering
{(a) Different Number of Trajectories with Five Obstacles in CPTO Framework} \vspace{1mm}\\
\begin{tabular}{cccc} 
\hline  
Trajectories &  Average Time   & Minimum Time & Maximum Time \\ 
\hline 
2 & 25.74 & 1.22 & 35.75 \\ 
3 & 32.08 & 30.68 & 39.33 \\ 
4 & 48.96 & 35.70 & 60.71 \\  
5 & 66.52 & 43.70 & 84.47 \\ \hline 
\end{tabular} 
\end{minipage}  
\begin{minipage}{0.45\textwidth}
\centering\vspace{2mm}
{(b) Different Number of Obstacles with Three Consensus Trajectories} \vspace{1mm}\\
\begin{tabular}{cccc} 
\hline  
Obstacles  &  Average Time   & Minimum Time & Maximum Time \\ 
\hline 
3 & 26.55 & 25.50 & 32.99 \\ 
4 & 30.52 & 28.98 & 44.08 \\ 
5 & 32.08 & 30.68 & 39.33 \\  
6 & 35.25 & 33.57 & 41.31 \\ \hline 
\end{tabular} 
\end{minipage}
\label{tab:combined_avg_time}
\end{table}
\subsection{Impact of Consensus Steps on Performance}
\label{subsec:tradeoff}  
{To further assess the impact of consensus steps within the CPTO framework, we conduct an ablation study with a simulation duration of $100\,\text{s}$, and the number of candidate trajectories $N_c=3$.} The perception setting for the EV follows the same configuration as described in Section~\ref{subsubsec:set_up}. We examine the performance of the EV with different consensus steps during a lane changing task under dense traffic flow and
perception uncertainties, as illustrated in Fig.~\ref{fig:Lane_changing_diagram}. Additionally, the BPHTO can adapt to such tasks, as it effectively handles lane merging conflict scenarios~\cite{zheng2024barrier}. In this context, SVs travel parallel to the centerline in each lane and are modeled using the IDM, with longitudinal speeds ranging from $8.5\,\text{m/s}$ to $18\,\text{m/s}$. The safe headway distance is set to $10\,\text{m}$, and the initial position of the EV is $[15\,\text{m}, 0\,\text{m}]^T$.
 
Table~\ref{tab:merge_results} shows that CPTO-8 and CPTO-15 achieve smaller mean absolute lateral jerk values and yaw rates than CPTO-0 and BPHTO, indicating better driving stability performance during lane changing. To obtain an intuitive view of this driving stability, Fig.~\ref{fig:Merge_comparasion} shows that the EV attempts to change lanes to the rear of SV1 from $50\,\text{m}$ to $150\,\text{m}$.  {
When Gap 1 (Fig.~\ref{fig:Lane_changing_diagram}) falls below the required headway due to the sudden deceleration of SV1, the EV aborts the lane change, returns to its original lane, and selects an alternative gap (Gap 3, Fig.~\ref{fig:Lane_changing_diagram}). This maneuver causes temporary heading angle fluctuations between $2.5\,\text{s}$ and $5\,\text{s}$. 
% During this process, longer consensus steps result in smaller fluctuations in the heading angle, leading to better driving consistency and stability.
{
Specifically, the nearly identical yaw rates of CPTO-15 ($0.0148\,\text{rad/s}$) and CPTO-8 ($0.0147\,\text{rad/s}$) indicate diminishing returns beyond 8 consensus steps, while still providing significant improvements over CPTO-0 and BPHTO. These results demonstrate that extended consensus planning enhances stability, with optimal benefits achieved at moderate step counts.}
Fig.~\ref{fig:snapshot_merge} clearly illustrates this entire process. These findings indicate that delayed transitions with shared consensus steps improve driving consistency in uncertain environments. Furthermore, all CPTO configurations maintain significantly greater longitudinal distances to the nearest SV than BPHTO, as shown by their mean absolute longitudinal spacing ($\mathcal{S}_{\text{lon, ma}}$): $13.3697\,\text{m}$ (CPTO-0), $13.6259\,\text{m}$ (CPTO-8), and $13.2591\,\text{m}$ (CPTO-15) versus $9.8486\,\text{m}$ (BPHTO).

% {All CPTO variants maintain 35-38\% greater safety margins than BPHTO, demonstrating robust collision avoidance regardless of configuration, with CPTO-8 showing the best balance between anticipation and reactivity.}}

In terms of computational efficiency, it can be seen that with the increasing number of consensus steps, the average optimization time slightly increases, from $34.25\,\text{ms}$ to $35.64\,\text{ms}$. The resulting $4\%$ increase in computation time from 0 to 15 consensus steps demonstrates the efficiency of our parallelized implementation. Notably, the increase in consensus steps does not significantly affect the overall computational efficiency, and the optimization can be performed in real time for each configuration. Furthermore, one can observe that with the increasing consensus steps from 8 to 15, the driving stability and safety performance exhibit no significant differences. This phenomenon suggests that increasing the consensus steps has a limited effect on overall performance.

%     \begin{figure}[tp]
%     \centering
%         \subfigure[]{
%             \label{fig:num_traj} \hspace{-4mm}
%         \includegraphics[scale=0.18]{figures/Optimization_time_boxplot_traj_num_refined.pdf}}\hspace{-2mm} 
%         \subfigure[]{
%             \label{fig:num_obs}
%         \includegraphics[scale=0.18]{figures/Optimization_time_boxplot_obs_num_refined.pdf}}\hspace{-2mm}
%     \vspace{-1mm}
%     \caption{{Computation time per cycle across varying numbers of obstacles and trajectories within the proposed CPTO framework. (a) Varying
%      number of trajectories. (b) Varying numbers of nearest considered SVs with three parallel trajectories.}}	
%     \label{fig:diss_computational_time}
%     \vspace{-1mm}
% \end{figure}	  
  
% Figure~\ref{fig:num_obs} presents a boxplot illustrating the distribution of optimization time as the considered number of obstacles increases from two to six. Accompanying this visualization,

Overall, our results highlight a critical trade-off between performance and computational efficiency in trajectory planning. While longer consensus steps may theoretically accommodate smoother trajectory adjustments, they do not universally guarantee superior performance. Excessive delays in transitioning to divergent segments may lead to abrupt corrections in the latter stages, increasing collision risks or violating kinematic limits. Conversely, overly conservative consensus steps risk unnecessary computational burden. To design an effective consensus segment, one should consider: the uncertainty inherent in environmental predictions, the ability to execute maneuvers in the divergent segment, and the real-time computational budget for online adaptation. Future research will focus on dynamic tuning of consensus steps, such as uncertainty-driven transitions or divergence thresholds, to achieve context-aware adaptability without sacrificing real-time feasibility. 

{
\subsection{Computational Time Analysis}
\label{subsec:comp}  
\label{subsec:computational_performance}
%-------------------------------------------------------
To evaluate the computational efficiency of the proposed CPTO framework, we conduct simulations on a cruise control task under a dense traffic scenario. Each simulation runs for $100\,\text{s}$, divided into 1000 discrete steps. Following the dense traffic setup in \cite{zheng2024}, we model the motion of SVs using the IDM model, with desired velocities ranging from $6\,\text{m/s}$ to $22\,\text{m/s}$. The target longitudinal cruise speed of the EV is set to $15\,\text{m/s}$.}
 
 {   
Table~\ref{tab:combined_avg_time} presents the computation times observed across various configurations of obstacles and candidate trajectories. We note that the average computation time increases with the number of trajectories and obstacles. However, the maximum optimization time stabilizes at approximately \(40\, \text{ms}\) once the number of considered obstacles exceeds four. This stabilization indicates that only the nearest SVs impact the optimization process in the consensus ADMM iteration. Remarkably, even in configurations involving high numbers of variables (up to 5175) and constraints (up to 7860), the CPTO framework supports real-time replanning while ensuring the safety of the EV in all configurations. Additionally, the EV remains safe in all configurations. Moreover, when optimizing two trajectories simultaneously, the minimum computation time recorded is just \(1.22\, \text{ms}\).  This rapid convergence demonstrates the efficiency of consensus ADMM in managing and concluding iterations promptly, thereby significantly reducing unnecessary computational overhead.} { Overall, these results demonstrate the scalability and efficiency of the CPTO framework in handling varying traffic densities without compromising computational performance or safety.} 
  
  % involving 40 timesteps, 11*3*5+40*6*5*2 + 3*6*5 +3*6*5+1*6*5= 2275 or 2275 +6*40*5*2 = 5175 variables, and 6*5+ 2*5+ 2*5 + 40*5*2 +40*6*5*2 + 40*6*5*2+6*3*5+6*3*5+6*5+6*40*5+6*40*5= 7860  constraints.
 
{
Furthermore, additional techniques can further enhance computational efficiency while ensuring safety in the CPTO framework. For instance, scenario pruning methods based on spatiotemporal reachability analysis \cite{bouzidi2025} and scenario reduction \cite{Ulfsjooon2022integratingpomdp} can improve scalability without sacrificing safety. Additionally, studies suggest that tree-structured branch elimination methods can leverage domain-specific expert knowledge to guide branching, effectively identifying potentially risky scenarios while maintaining safety guarantees \cite{zhanglu2020efficientuncertainty}. Investigating these strategies remains an area for future research.}

\section{Conclusions}
	\label{sec:con} 
This paper presents a novel CPTO approach for real-time, consistent, and safe trajectory planning for autonomous driving in partially observed environments. The CPTO framework introduces a consensus safety barrier module, ensuring that each generated trajectory maintains a consistent and safe segment, even when faced with varying levels of obstacle detection accuracy. By transforming the complex non-convex trajectory planning problem into a series of manageable low-dimensional QP subproblems and leveraging parallel consensus optimization, the proposed framework enables large-scale optimization in real time.
Extensive experiments demonstrate that our approach outperforms several state-of-the-art methods in terms of safety and task accuracy. We also investigate the influence of consensus steps in dense traffic environments, revealing a trade-off between task performance and computational efficiency. While the CPTO approach shows great promise, the assumption regarding obstacle detection thresholds may not be feasible in all situations, particularly under low visibility or sensor failure conditions such as occlusion by large trucks or buildings in urban intersections. One potential solution is to leverage reachability analysis~\cite{park2023occlusion} for risk assessment and incorporate it as state constraints within the consensus ADMM iteration framework. As part of our future work, we will focus on exploring adaptive consensus strategies to further enhance the system's robustness in these challenging environments. \vspace{-5mm}
{\appendix[Derivation of Consensus ADMM Iterations]
In the following, we provide the derivation of analytical solutions for the primal variables $\mathbf{C}_{\theta}$, $\mathbf{C}_{x}$, and $\mathbf{C}_{y}$ during ADMM iteration in Section~\ref{subsubsec:primal_uodate}. 
 
Derivation for the primal variable $\mathbf{C}_{\theta}$: 
Leveraging the polar transformation \eqref{eq:angle_value}, the subproblem \eqref{eq:sub_qp1} can be converted into the following constrained least squares problem: 
\begin{alignat}{2} \vspace{-0mm}
    \mathbf{C}^{\iota+1}_{\theta}  &:= \displaystyle\operatorname*{ \text{argmin}}_{\mathbf{C}_{\theta}}~~
    \mathcal{L}   \Big(  \{\mathbf{C}_\theta\}, \{ \mathbf{C}^{\iota}_x, \mathbf{Z}^{\iota}_x\}, \{ \mathbf{C}^{\iota}_y, \mathbf{Z}^{\iota}_y\},  \{\bm{\omega}^{\iota}\}, \{ \mathbf{D}^{\iota}\}, \nonumber \\
   &\qquad\qquad\qquad \{ \mathbf{Y}^{\iota}_x, \mathbf{Y}^{\iota}_y, \mathbf{Y}^{\iota}_{\theta}\}, \{ \bm{\lambda}^{\iota}_{\text{cons},x}, \bm{\lambda}^{\iota}_{\text{cons},y}, \bm{\lambda}^{\iota}_{\text{cons},\theta} \}, \nonumber \\ 
    &\qquad\qquad\qquad \{\bm{\lambda}^{\iota}_{\theta},  \bm{\lambda}^{\iota}_{x}, \bm{\lambda}^{\iota}_{y},  \bm{\lambda}^{\iota}_{\text{obs},x}, \bm{\lambda}^{\iota}_{\text{obs},y} \}   \Big) \nonumber\\
   & = \displaystyle\operatorname*{min}_{\mathbf{C}_{\theta}}~~\frac{1}{2} \mathbf{C}^T_\theta Q_{\theta}\mathbf{C}_\theta  \notag\\
& \qquad\qquad +  \bm{\lambda}_{\theta}^{\iota T}
 \left( \mathbf{W}^T_B\mathbf{C}_{\theta} -  
 \arctan \left(\frac{\dot{\mathbf{W}}^T_B  \mathbf{C}^{\iota}_{y} }{ \dot{\mathbf{W}}^T_B \mathbf{C}^{\iota}_{x} }\right) \right) \notag
  \\
   &\qquad\qquad +  \bm{\lambda}_{\text{cons},\theta}^{\iota T} (\mathbf{A}^T_{\text{cons},\theta}  \mathbf{C}_{\theta}  -\mathbf{Y}^{\iota}_{\theta} )  \notag\\ 
&\qquad\qquad + \frac{{\rho}_{\theta}}{2} \left\| \mathbf{W}^T_B\mathbf{C}_{\theta} -  
 \arctan \left(\frac{ \dot{\mathbf{W}}^T_B \mathbf{C}^{\iota}_{y} }{\dot{\mathbf{W}}^T_B  \mathbf{C}^{\iota}_{x} }\right) \right\| ^2  \notag\\
 &\qquad\qquad + \frac{\rho_{\text{cons},\theta}}{2} \left\|\mathbf{A}^T_{\text{cons},\theta}  \mathbf{C}_{\theta}  -\mathbf{Y}^{\iota}_{\theta} \right\| ^2
 \notag\\ 
   &   \qquad \text{s.t.}\quad
    [\mathbf{A}_0 \quad \mathbf{A}_{f,\theta} ]^T \mathbf{C}_{\theta}  = [\bm{\theta}_0\quad \dot{\bm{\theta}}_0\quad \mathbf{0}]^T.  \nonumber \vspace{-5mm}
 \end{alignat}

\[
\bm{\lambda}^{\iota+1}_x
=
\bm{\lambda}^{\iota}_x
+
\rho_x
\mathbf G^T
(
\mathbf G\mathbf C_x^{\iota+1}
-\mathbf F_x
+\mathbf Z_x^{\iota+1}
).
\]

As a result, we can obtain the following analytical solutions for the variable $\mathbf{C}_{\theta}$: 
\begin{equation} \vspace{-1mm}
   \mathbf{C}_{\theta} = \mathbf{\Xi}_{\theta}^\dagger \mathbf{b}_{\theta},
   \notag
 \vspace{-1mm} \end{equation} 
where 
\( \mathbf{\Xi}_{\theta}^\dagger \) denotes the pseudoinverse of \( \mathbf{\Xi}_{\theta} \), 
\[ \bm{\Xi}_{\theta}
= \begin{bmatrix} \mathbf{Q}_{\theta} + \rho_{\theta}\mathbf{W}_B\mathbf{W}^T_B + \rho_{\text{cons},\theta}\mathbf{A}_{\text{cons},\theta}\mathbf{A}^T_{\text{cons},\theta}\\ \mathbf{A}_0 \\ \mathbf{A}_{f,\theta} \end{bmatrix}, \]
% \[ \mathbf{b}_{\theta} = [ \mathbf{b}_{\theta,0} \quad [\bm{\theta}_0\quad \dot{\bm{\theta}}_0]^T  \quad \mathbf{0} ]^T. \] 
\[ \mathbf{b}_{\theta} = \begin{bmatrix} \mathbf{b}_{\theta,0} \\ [\bm{\theta}_0\quad \dot{\bm{\theta}}_0]^T \\ \mathbf{0} \end{bmatrix}. \] 
Here, $\mathbf{b}_{\theta,0}$ is given by:
\begin{alignat}{2}\vspace{-0mm} \mathbf{b}_{\theta,0} =  
&- \mathbf{W}_B \bm{\lambda}_{\theta}^{\iota}  + \rho_{\theta}\mathbf{W}_B\arctan \left(\frac{ \dot{\mathbf{W}}^T_B \mathbf{C}^{\iota}_{y} }{ \dot{\mathbf{W}}^T_B \mathbf{C}^{\iota}_{x} }\right) \nonumber\\
&- \mathbf{A}_{\text{cons},\theta} \bm{\lambda}_{\text{cons},\theta}^{\iota} +\rho_{\text{cons},\theta}\mathbf{A}_{\text{cons},\theta} \mathbf{Y}^{\iota}_{\theta}. \nonumber \vspace{-1mm}
\end{alignat} 

Derivation for the primal variable $\mathbf{C}_x$: 
We aim to solve the following constrained least squares problem:   
\begin{alignat}{2} \vspace{-0mm} 
    \mathbf{C}^{\iota+1}_{x} &:= \displaystyle\operatorname*{ \text{argmin}}_{\mathbf{C}_{x}}~~
    \mathcal{L}   \Big(  \{\mathbf{C}^{\iota}_\theta\}, \{ \mathbf{C}_x, \mathbf{Z}^{\iota}_x\}, \{ \mathbf{C}^{\iota}_y, \mathbf{Z}^{\iota}_y\},  \{\bm{\omega}^{\iota}\}, \{ \mathbf{D}^{\iota}\}, \nonumber \\
   &\qquad\qquad\qquad \{ \mathbf{Y}^{\iota}_x, \mathbf{Y}^{\iota}_y, \mathbf{Y}^{\iota}_{\theta}\}, \{ \bm{\lambda}^{\iota}_{\text{cons},x}, \bm{\lambda}^{\iota}_{\text{cons},y}, \bm{\lambda}^{\iota}_{\text{cons},\theta} \}, \nonumber \\ 
    &\qquad\qquad\qquad \{\bm{\lambda}^{\iota}_{\theta},  \bm{\lambda}^{\iota}_{x}, \bm{\lambda}^{\iota}_{y},  \bm{\lambda}^{\iota}_{\text{obs},x}, \bm{\lambda}^{\iota}_{\text{obs},y} \}   \Big) \notag\\ 
      & = \displaystyle\operatorname*{min}_{\mathbf{C}_{x}}~~\frac{1}{2} \mathbf{C}^T_x Q_{x}\mathbf{C}_x  
      + \bm{\lambda}_{\theta}^{\iota T}   ( \dot{\mathbf{W}}^T_B \mathbf{C}_{x}  - \mathbf{V} \cdot \cos{ (\mathbf{W}^T_B\mathbf{C}^{\iota}_{\theta})}  )   \notag\\
 &\qquad\qquad  + \bm{\lambda}_{\text{obs},x}^{\iota T}  \left(\mathbf{A}_h \mathbf{C}_{x}  - \mathbf{O}_{x}  - \mathbf{L}_x \cdot  \mathbf{D}^{\iota}  \cdot  \cos ( \bm{\omega}^{\iota})\right)  \notag\\ 
   &\qquad\qquad +  \bm{\lambda}_{\text{cons},x}^{\iota T} (\mathbf{A}^T_{\text{cons},x}  \mathbf{C}_{x}  -\mathbf{Y}^{\iota}_{x} ) + \bm{\lambda}_{x}^{\iota T} \mathbf{C}_{x}  \notag\\ 
 &\qquad\qquad  + \frac{\rho_{\theta}}{2}  \left\| \dot{\mathbf{W}}^T_B \mathbf{C}_{x}  - \mathbf{V} \cdot \cos{ (\mathbf{W}^T_B\mathbf{C}^{\iota}_{\theta})} \right\|_{2}^{2} 
   \notag \\
&\qquad\qquad +\frac{\rho_{\text{obs}}}{2} \left\| \mathbf{A}_h \mathbf{C}_{x}  - \mathbf{O}_{x}  - \mathbf{L}_x \cdot  \mathbf{D}^{\iota}  \cdot  \cos ( \bm{\omega}^{\iota})  \right\|_{2}^{2} \notag\\
  &\qquad\qquad + \frac{\rho_{\text{cons},x}}{2}  \left\| \mathbf{A}^T_{\text{cons},x}  \mathbf{C}_{x}  -\mathbf{Y}^{\iota}_{x} \right\|_{2}^{2}
 \notag \\
&\qquad\qquad + \frac{\rho_x}{2} \left\| \mathbf{G} \mathbf{C}_x  -\mathbf{F}_x + \mathbf{Z}^{\iota}_x \right\|_{2}^{2}\notag \\ \vspace{-0mm} 
   &   \qquad \text{s.t.}\quad
   [ \mathbf{A}_0 \quad \mathbf{A}_{f,x} ]^T \mathbf{C}_{x}  = [\mathbf{P}_{x,0}\quad \dot{\mathbf{P}}_{x,0}\quad \mathbf{P}_{x,N} ]^T.    \nonumber 
\vspace{-5mm} \end{alignat}
 
As a result, we can get the following analytical solutions for the variable $\mathbf{C}_{x}$: 
\begin{equation} \vspace{-0mm}
   \mathbf{C}^{\iota+1}_{x} = \mathbf{\Xi}_{x}^\dagger \mathbf{b}_{x},
\notag \vspace{-0mm} \end{equation} 
where \( \mathbf{\Xi}_{x}^\dagger \) denotes the pseudoinverse of \( \mathbf{\Xi}_{x} \),
\[ \bm{\Xi}_{x} = \begin{bmatrix}   \bm{\Xi}_{x,0} 
\\ \mathbf{A}_0\\ \mathbf{A}_{f,x} \end{bmatrix}, \quad  \mathbf{b}_{x} = \begin{bmatrix} 
  \mathbf{b}_{x,0}
  \\ [\mathbf{P}_{x,0}\quad \mathbf{V}_{x,0} ]^T \\ \mathbf{P}_{x,g}\end{bmatrix}, \] 
and $\mathbf{\Xi}_{x,0} $ and $\mathbf{b}_{x,0} $ are given by:
\begin{alignat}{2} \vspace{-0mm}
\mathbf{\Xi}_{x,0} = & \mathbf{Q}_{x} + \rho_{\theta}\dot{\mathbf{W}}_B\dot{\mathbf{W}}^T_B 
+ \rho_{\text{obs}}\mathbf{A}^T_h  \mathbf{A}_h\nonumber \\
    & + \rho_{\text{cons},x}\mathbf{A}_{\text{cons},x}\mathbf{A}^T_{\text{cons},x}
+ \rho_{x}  \mathbf{G}^T  \mathbf{G} ,\notag 
\vspace{-0mm} \end{alignat}    
\begin{alignat}{2} \vspace{-0mm}
 \mathbf{b}_{x,0} = & -\bm{\lambda}_{x}^{\iota}
-\dot{\mathbf{W}}_B\bm{\lambda}_{\theta}^{\iota} 
- \mathbf{A}^T_h\bm{\lambda}_{\text{obs},x}^{\iota } \nonumber \\
    & +\rho_{\theta}\dot{\mathbf{W}}_B \mathbf{V} \cdot \cos{( \mathbf{W}^T_B\mathbf{C}^{\iota}_{\theta}} ) 
 \notag\\
& + \rho_{\text{obs}}\mathbf{A}^T_h ( \mathbf{O}_{x}  + \mathbf{L}_x \cdot  \mathbf{D}^{\iota} \cdot  \cos{(\bm{\omega}^{\iota})} ) \notag\\
 & + \rho_{\text{cons},x}\mathbf{A}_{\text{cons},x}  \mathbf{Y}^{\iota}_{x}   + \frac{\rho_x}{2} \mathbf{G}^T (\mathbf{F}_x-\mathbf{Z}^{\iota}_x )
 .\notag 
\vspace{-5mm} \end{alignat}

Similarly, we can obtain the following iteration result for variable $ \mathbf{C}_y$ : 
\begin{equation} \vspace{-0mm}
   \mathbf{C}^{\iota+1}_{y} = \mathbf{\Xi}_{y}^\dagger \mathbf{b}_{y},
   \notag
 \vspace{-0mm} \end{equation} 
where \( \mathbf{\Xi}_{y}^\dagger \) denotes the pseudoinverse of \( \mathbf{\Xi}_{y} \),
\[ \bm{\Xi}_{y} ={ \begin{bmatrix}   \bm{\Xi}_{y,0} 
\\ \mathbf{A}_{0} \\ \mathbf{A}_{f,y} \end{bmatrix}, }  \quad \mathbf{b}_{y} = \begin{bmatrix} 
  \mathbf{b}_{y,0}
  \\ [\mathbf{P}_{y,0}\quad \mathbf{V}_{y,0} ]^T \\ \mathbf{P}_{y,g}\end{bmatrix}, \] 
and $\mathbf{\Xi}_{y,0} $ and $\mathbf{b}_{y,0} $ are define as:
\begin{alignat}{2} \vspace{-0mm}
\mathbf{\Xi}_{y,0} = & \mathbf{Q}_{y} + \rho_{\theta}\dot{\mathbf{W}}_B\dot{\mathbf{W}}^T_B 
+ \rho_{\text{obs}}\mathbf{A}^T_h  \mathbf{A}_h\nonumber \\
    & + \rho_{\text{cons},y}\mathbf{A}_{\text{cons},y}\mathbf{A}^T_{\text{cons},y}
+ \rho_{y}  \mathbf{G}^T  \mathbf{G} ,\notag 
\vspace{-0mm} \end{alignat}    
\begin{alignat}{2} \vspace{-0mm} 
 \mathbf{b}_{y,0} = & -\bm{\lambda}_{y}^{\iota}
-\dot{\mathbf{W}}_B\bm{\lambda}_{\theta}^{\iota} 
- \mathbf{A}^T_h\bm{\lambda}_{\text{obs},y}^{\iota }  \notag\\
& +\rho_{\theta}\dot{\mathbf{W}}_B \mathbf{V} \cdot \sin{ (\mathbf{W}^T_B\mathbf{C}^{\iota}_{\theta})}  
 \notag\\
& + \rho_{\text{obs}}\mathbf{A}^T_h ( \mathbf{O}_{y}  + \mathbf{L}_y \cdot  \mathbf{D}^{\iota} \cdot  \sin{(\bm{\omega}^{\iota})} ) \notag\\
 & + \rho_{\text{cons},y}\mathbf{A}_{\text{cons},y}  \mathbf{Y}^{\iota}_{y}   + \frac{\rho_y}{2} \mathbf{G}^T (\mathbf{F}_y-\mathbf{Z}^{\iota}_y ) .\notag 
\vspace{-0mm} \end{alignat}

\bibliographystyle{IEEEtran}
\bibliography{egbib}

\vfill

\end{document}